\documentclass[10pt,twocolumn,letterpaper]{article}

\usepackage{btas}
\usepackage{times}
\usepackage{epsfig}
\usepackage{amsmath,amssymb,amsfonts}
\usepackage{algorithmic}
\usepackage{graphicx}
\usepackage{caption}
\usepackage{subfig}
\usepackage{textcomp}
\usepackage{times}
\usepackage{epsfig}
\usepackage{dsfont}
\usepackage{calrsfs}
\usepackage{threeparttable}
\usepackage[linesnumbered,ruled,vlined]{algorithm2e}
\usepackage{mathtools}
\usepackage{array}
\usepackage{booktabs}
\usepackage{bm} 
\usepackage{multirow}
\usepackage{enumerate}
\usepackage{enumitem}
\usepackage{makecell}
\usepackage[subnum]{cases}
\usepackage{dsfont}
\graphicspath{ {FIGURES/} }

\usepackage{balance}
\usepackage[pagebackref=false,breaklinks=true,letterpaper=true,colorlinks,bookmarks=false]{hyperref}
 \newcommand\Myperm[2][^n]{\prescript{#1\mkern-2.5mu}{}P_{#2}} 
 
\newcommand\tab[1][0.5cm]{\hspace*{#1}} 

\btasfinalcopy 

\makeatletter

\makeatletter 
\def\ps@IEEEtitlepagestyle{
 \def\@oddfoot{\mycopyrightnotice}
 \def\@evenfoot{}
}
\def\mycopyrightnotice{
 {\hfill \footnotesize 978-1-7281-1522-1/19/\$31.00 \copyright 2019 IEEE\hfill}
}\makeatother

\begin{document}

\title{Smartphone Camera De-identification while Preserving Biometric Utility}
\author{Sudipta Banerjee and Arun Ross \\
Michigan State University \\
{\tt\small \{banerj24, rossarun\} @cse.msu.edu}
}

\maketitle

\setcounter{footnote}{0}
\thispagestyle{empty}

\begin{abstract}
The principle of Photo Response Non Uniformity (PRNU) is often exploited to deduce the identity of the smartphone device whose camera or sensor was used to acquire a certain image. In this work, we design an algorithm that perturbs a face image acquired using a smartphone camera such that (a) sensor-specific details pertaining to the smartphone camera are suppressed (sensor anonymization); (b) the sensor pattern of a different device is incorporated (sensor spoofing); and (c) biometric matching using the perturbed image is not affected (biometric utility). We employ a simple approach utilizing Discrete Cosine Transform to achieve the aforementioned objectives. Experiments conducted on the MICHE-I and OULU-NPU datasets, which contain periocular and facial data acquired using 12 smartphone cameras, demonstrate the efficacy of the proposed de-identification algorithm on three different PRNU-based sensor identification schemes. This work has application in sensor forensics and personal privacy. 

\end{abstract}

{\let\thefootnote\relax\footnotetext{\mycopyrightnotice}}

\section{Introduction}
\label{sec:intro}
Sensor identification, or source attribution,~\cite{Bayram_ICIP_05} refers to the automated deduction of sensor identity from a digital image. Photo Response Non-Uniformity (PRNU) based sensor identification algorithms have been successfully used in the context of RGB~\cite{Lukas_TIFS_06, UniquePRNU, PRNU_Robustness_2009} and near-infrared (NIR)~\cite{ Kalka_CVPRW_15, Uhl5_IJCB_17, Vatsa_18} images. PRNU based sensor identification algorithms typically use image denoising techniques to elicit sensor specific details from an image. The forensic utility of these algorithms has been well established in the literature~\cite{Galdi_PRL_15}. 

Since smartphone devices are intricately linked to their owners, sensor identification using images from smartphone cameras can inevitably lead to person identification. This poses privacy concerns to the general populace~\cite{youtube} and, especially, to photojournalists~\cite{AnonRef_2011}. \textit{Sensor de-identification} can mitigate such concerns by removing sensor specific traces from the image. A number of sensor de-identification algorithms, particularly in the context of PRNU suppression, have been developed in the literature~\cite{PRNU_Perturb1_2014, PRNU_attack2}. PRNU suppression can be done by either PRNU anonymization or PRNU spoofing. 

\begin{figure}[]
\centering

    \includegraphics[scale=.37]{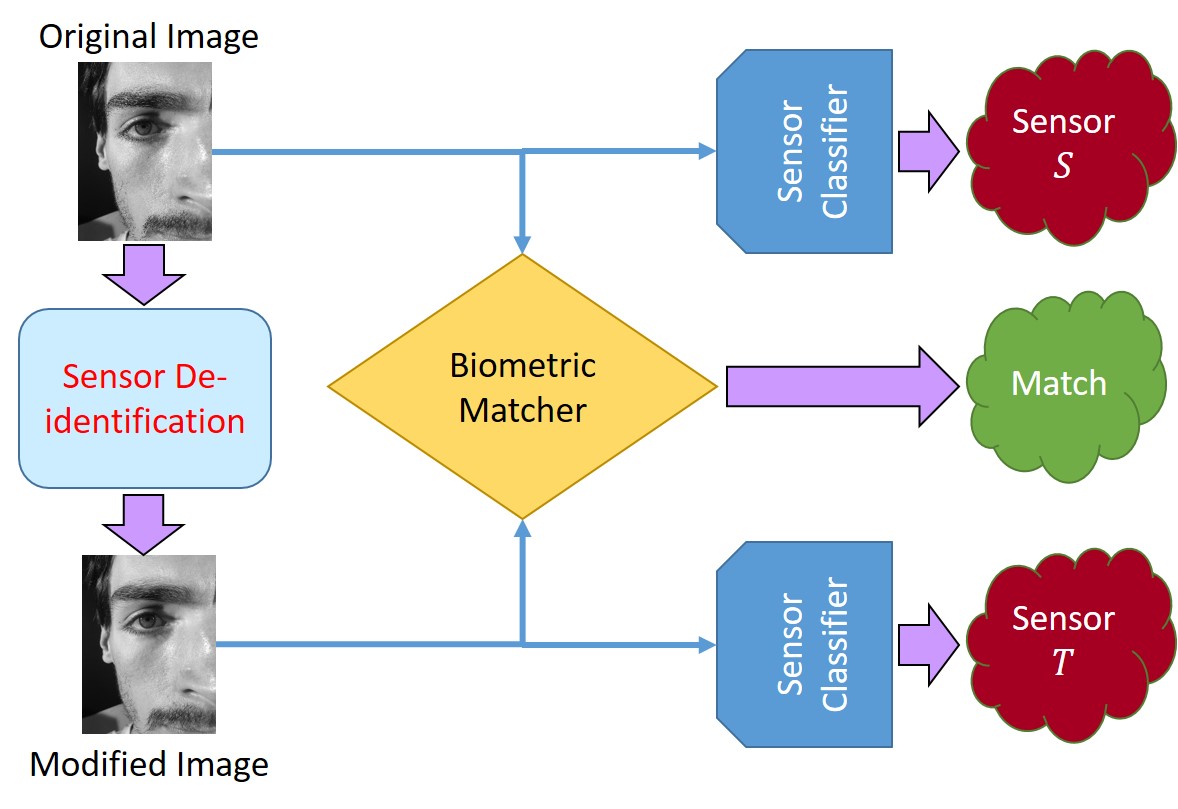} 

\caption{The objective of our work. The original biometric image is modified such that the {\em sensor classifier} associates it with a different sensor, while the {\em biometric matcher} successfully matches the original image with the modified image.}
\label{Fig: Obj}
\end{figure}

PRNU anonymization is typically accomplished using strong filtering schemes~\cite{PRNU_attack3_2014} that perturb the PRNU pattern, or irreversible transformations such as `seam carving'~\cite{PRNU_Perturb1_2014} that systematically remove rows and columns from the image. This produces geometrical perturbations which impair correct source attribution. In~\cite{PRNU_attack5_2010}, the authors performed `Signature Removal' by subtracting a scaled version of the PRNU signature (sensor reference pattern) from an image. For an original image, $\bm{I}$, belonging to sensor $S$, the PRNU anonynimized image is obtained as $\bm{I'}=\bm{I} - \gamma \bm{\hat{K}_{S}}$. Here, $\gamma$ is the scaling factor for the sensor reference pattern $\bm{\hat{K}_{S}}$. PRNU spoofing, on the other hand, involves deliberately confounding the image such that a sensor classifier attributes it to a different {\em target} sensor rather than the actual {\em source} sensor. This can be achieved by injecting the PRNU pattern of the target sensor directly into the test image~\cite{Ref_Fingerprintcopy2}. The modified image becomes $\bm{I'}=[\bm{I}+\bm{I}\times\gamma \bm{\hat{K}_{T}}]$. Here, $\bm{I}$ is the original image and $\bm{\hat{K}_{T}}$ is the reference pattern of the target sensor $T$. The term $\gamma$ is a scalar parameter that needs to be empirically tuned to achieve the spoofing. Alternatively, one can first subtract the reference pattern or the `fingerprint' of the source sensor and then insert the target sensor pattern; this is known as `Signature Substitution'~\cite{PRNU_attack5_2010}. The modified image is represented as $\bm{I'}=\bm{I} - \gamma \bm{\hat{K}_{S}} + \beta \bm{\hat{K}_{T}}$. $\bm{I}$ belongs to the source sensor $S$, whose reference pattern is $\bm{\hat{K}_{S}}$. The terms $\gamma$ and $\beta$ are scalars which can be optimized through grid-line search. 

Majority of the work discussed above focus on images depicting natural scenes. However, for biometric images (face, iris, fingerprints), the perturbations may degrade the quality of the samples making them unsuitable for person recognition. Limited work has been done in the context of sensor de-identification for biometric samples~\cite{Uhl4_ICB_12}. In~\cite{Ross_19}, the authors proposed an iterative perturbation algorithm, which applied patch-wise modifications to the input biometric image to perform sensor de-identification while retaining its biometric utility. Current methods utilize parameters optimized through exhaustive search techniques and require computation of sensor reference patterns to achieve sensor de-identification~\cite{Ref_Fingerprintcopy2, PRNU_attack5_2010, Uhl4_ICB_12}. Semi-adversarial learning~\cite{SAN} based methods can be another viable way to accomplish sensor de-identification while retaining biometric matching utility. However, such methods are complex and require a large number of training images.~\textbf{The objective of our work is to develop a rather simple method to perform sensor de-identification, while preserving the biometric recognition utility of the images.} The key idea is illustrated in Figure~\ref{Fig: Obj}. The merits of the proposed method are as follows.\\

\noindent 1. Designing a sensor de-identification algorithm that can perform both PRNU anonymization and PRNU spoofing in a non-iterative fashion. This addresses the computational overhead incurred by the algorithms in~\cite{PRNU_attack3_2014, Ross_19}.\\
2. The proposed de-identification algorithm is applicable to different PRNU estimation schemes and works irrespective of the source and target sensors. This eliminates the need for parameter optimization and computation of the reference patterns corresponding to each pair of source and target sensors as required in~\cite{Ref_Fingerprintcopy2, PRNU_attack5_2010}.\\  
3. The proposed algorithm causes minimal degradation to the biometric content of the images, thus retaining their biometric utility. 

The remainder of the paper is organized as follows. Section~\ref{sec:prnu} discusses the concept of PRNU that has been used for sensor identification from images. Section~\ref{sec:prop} describes the proposed algorithm for PRNU anonymization and PRNU spoofing. In section~\ref{sec:expts}, we describe the dataset, the experimental protocol and the results. Section~\ref{sec:concl} concludes the paper.

\section{Photo Response Non-Uniformity (PRNU)}
\label{sec:prnu}

Photo Response Non-Uniformity (PRNU)~\cite{Lukas_TIFS_06} manifests due to anomalies present in the silicon wafer used during sensor fabrication. PRNU is a type of sensor pattern noise which arises due to the variation in the response of the pixels across the sensor plane to the same light intensity and is intrinsically linked to each sensor. Therefore, PRNU can be used for sensor identification. Sensor identification using PRNU requires the computation of the \textit{sensor reference pattern} from a set of \textit{training} images acquired using a specific sensor. When an unknown test image is provided, the {\em noise residual} of that image is computed and correlated with the reference pattern of several different sensors. The test image is attributed to the sensor whose reference pattern yields the highest correlation value. 

While a number of algorithms have been proposed in the literature to perform sensor identification~\cite{Review_ref1, Review_ref2}, in this work we will focus on 3 of them~\textemdash~(i) Enhanced PRNU Estimation, (ii) Maximum Likelihood based PRNU Estimation and (iii) Phase PRNU Estimation methods \textemdash ~as these 3 fundamental PRNU based methods have been widely used in the literature. Each method generates the sensor reference pattern from a set of training images as described below. 

\noindent  \textbf{1. Enhanced PRNU Estimation:} The reference pattern is computed as an average of the noise residuals corresponding to the training images~\cite{Li_TIFS_10}.\\
 \textbf{2. Maximum Likelihood based PRNU Estimation (MLE PRNU):} The noise residuals from the training set are weighted by the images and normalized by the sum of the squares of the pixel intensity values of the training images~\cite{Lukas_TIFS_08}, and furthermore, subjected to Wiener filtering and zero-mean operations to remove interpolation artifacts. \\
 \textbf{3. Phase PRNU Estimation:} The phase component of the PRNU is computed in the Fourier domain and then averaged across the training images~\cite{Kang_TIFS_12} to generate the reference pattern.

The test noise residual can be extracted by either simply applying a wavelet based denoising filter to the test image or by applying an enhancement model to further attenuate scene influences, as done in Enhanced PRNU Estimation~\cite{Li_TIFS_10}.

\section{Proposed Method}
\label{sec:prop}

\begin{figure}[h]
\centering

    \includegraphics[scale=.45]{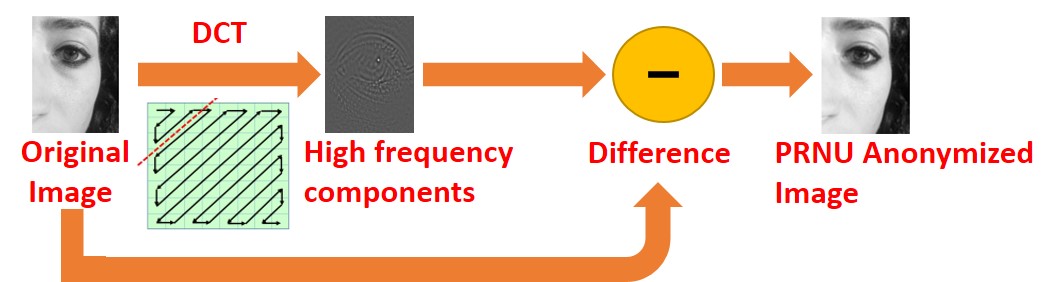} 

\caption{Illustration of PRNU Anonymization. The DCT coefficients are arranged such that the top-left portion has the low frequency components while the bottom-right portion encapsulates the high frequency information. The PRNU anonymized image is the result of suppression of high frequency components (see Algorithm~\ref{alg:Anon}, here $\eta =0.9$).}
\label{Fig: PRNUAnon}
\end{figure} 

\begin{algorithm}[t]
\scriptsize
  \caption{\label{alg:Anon}PRNU Anonymization.}
  \KwIn{An image $\bm{I}$ of size $h\times w$ and parameter $\eta$}
  \KwOut{PRNU anonymized image $\bm{I'}$}

 Apply 2-dimensional DCT to $\bm{I}$ \hspace{6cm} $\bm{I}_{dct} = DCT(\bm{I})$
 
 Compute $m = \min(h,w)$ and $\alpha = round(\eta\times m)$
 
 Extract the high frequency components as follows: $\bm{I}_{high} = High(\bm{I}_{dct},\alpha)$, \hspace{4.5cm} where, the $High(\cdot,\cdot)$ operator extracts the lower triangular portion of the DCT coefficients along the anti-diagonal direction, regulated by $\alpha$
 
 Extract the low frequency components as follows: $\bm{I}_{low} = \bm{I}_{dct} -\bm{I}_{high}$
 
 Apply inverse DCT to obtain the modified image $\bm{I'}= DCT^{-1}(\bm{I}_{low})$ 
 
 Return the modified image $\bm{I'}$
     
\end{algorithm}

\begin{figure}
\centering

    \includegraphics[scale=.45]{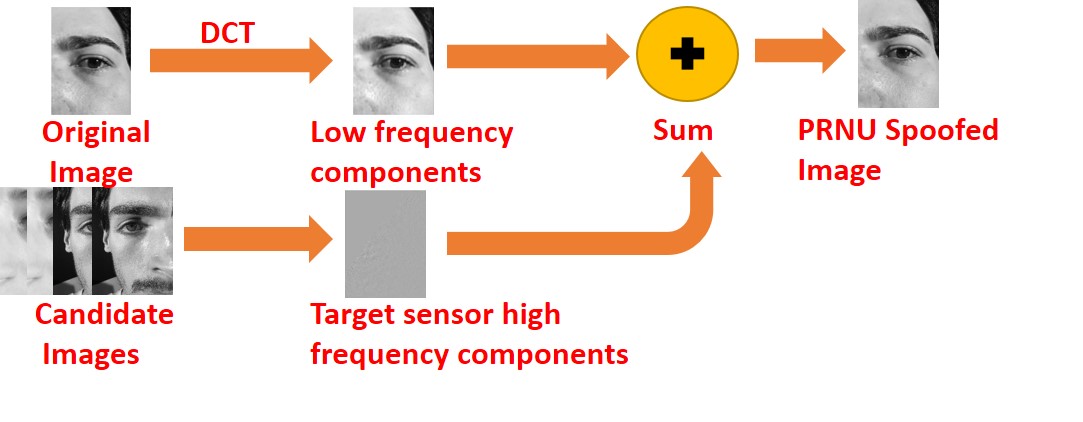} 

\caption{Illustration of PRNU Spoofing. The high frequency components in the original image are suppressed first, the residue being the low frequency components. The high frequency components of the target sensor are further computed from the candidate images, and added to the low frequency components of the original image, resulting in the PRNU spoofed image (see Algorithm~\ref{alg:Spoof}, here $\eta =0.7$).}
\label{Fig: PRNUSpoof}
\end{figure}

Discrete Cosine Transform (DCT) has been successfully used for lossy image compression~\cite{DCT} or for improving source camera identification~\cite{DCT_PRNU}. The coefficients located in the top-left portion capture the low frequency components while the bottom-right coefficients encode the high frequency components. Our goal is to modify the images to perturb the PRNU pattern resulting in sensor de-identification. PRNU is a noise-like component which is dominated by the high frequency components present in an image. Thus, we propose to transform the image into the DCT domain and modulate the DCT coefficients such that the high frequency components are suppressed, while retaining the low frequency components. By suppressing the high frequency components, we mask the sensor pattern present in the image. On the other hand, we retain the low frequency components which primarily contain the scene details in the image. The scene details are pivotal for biometric recognition. Thus, we ensure preservation of the biometric utility of the image. We then apply the inverse DCT, and the output is the modified image. 

\begin{algorithm}[h]
\scriptsize
  \caption{\label{alg:Spoof}PRNU Spoofing.}
  \KwIn{An image $\bm{I}$ of size $h\times w$ belonging to source sensor $S$, a set of $N$ candidate images belonging to the target sensor $T$, where each image of size $p\times q$ is denoted as $\bm{G}^i$ ($i=[1,\cdots,N]$) and $\eta$ }
  \KwOut{PRNU spoofed image $\bm{I'}$} 
 Set $i=1$ 
  
 Apply 2-dimensional DCT to $\bm{I}$, $\bm{I}_{dct} = DCT(\bm{I})$
 
 Extract the low frequency and high frequency components, $\bm{I}_{low}$ and $\bm{I}_{high}$ as described in Algorithm~\ref{alg:Anon} and set $\bm{I}_{high}=0$
 
 Compute $\alpha = round(\eta\times \min(p, q))$
 
 \Repeat{$i=N$}{
 \nl -- Apply 2-dimensional DCT to $\bm{G}^i$, $\bm{G}^i_{dct} = DCT(\bm{G}^i)$
 
 \nl -- Extract the high frequency components as follows: $\bm{G}^{i}_{high} = High(\bm{G}^i_{dct},\alpha)$
  
 \nl -- Apply inverse DCT to the high frequency content as follows: $\bm{G'}_i= DCT^{-1}(\bm{G}^{i}_{high})$ 
 
 \nl -- Add the images to generate $\bm{T}_{high} += \bm{G'}_i$ and increment $i += 1$
  
 } 
 
 Divide by the number of images $\bm{T}_{high} = \frac{\bm{T}_{high}}{N}$ 
 
 Resize $\bm{T}_{high}$ to $h\times w$ using bicubic interpolation
 
 Apply inverse DCT to obtain the modified image $\bm{I'}= DCT^{-1}(\bm{T}_{high}+\bm{I}_{low})$ 
 
 Return the modified image $\bm{I'}$
    
\end{algorithm}

To achieve sensor de-identification we perform both (i) PRNU Anonymization and (ii) PRNU Spoofing. 

\noindent \textbf{PRNU Anonymization:} Given an image $\bm{I}$, we first subject it to DCT to yield $\bm{I_{dct}}$. We intend to suppress the high frequency information without impairing the low frequency details. To achieve this goal, we define a parameter $\alpha$ that serves as a regulator for high frequency suppression. $\alpha$ is computed as the product of the minimum of the height and width of the image $min(h,w)$, and a user-defined parameter $\eta$, rounded off to the nearest integer. All DCT coefficients present in the interval $[row=\alpha : h, col=\alpha : w]$ are set to zero. Thus, $\alpha$ represents the threshold for the suppression of the DCT coefficients, and that threshold is a function of the image dimensions. We discard the high frequency components and then apply inverse DCT which results in the PRNU anonymized image $\bm{I'}$. The steps are described in Algorithm~\ref{alg:Anon}. The process of PRNU anonymization is illustrated using an example image in Figure~\ref{Fig: PRNUAnon}.

\begin{table*}[h]
\centering
\caption{Dataset specifications. The top block corresponds to MICHE-I dataset~\cite{MICHE} and the bottom block corresponds to OULU-NPU face dataset~\cite{Oulu_1}. In the MICHE-I dataset, we denote the brand Apple as `Device 1'  and the brand Samsung as `Device 2'. Two different smartphones belonging to the same brand and model, \textit{e.g.}, Apple iPhone5, are distinguished as `UNIT I' and `UNIT II'.}
\label{Tab: Dataset}
\scalebox{0.79}{
\begin{tabular}{|cccccc|} \hline
\multicolumn{1}{|c}{\textbf{Smartphone Brand and Model}}                                         & \textbf{Device Identifier} & \textbf{Sensor}    & \textbf{Image Size} & \multicolumn{1}{c}{\begin{tabular}[c]{@{}c@{}}\textbf{Number of Images/}\\  \textbf{Number of Subjects}\\ \textbf{(Training Set)}\end{tabular}} & \multicolumn{1}{c|}{\begin{tabular}[c]{@{}c@{}}\textbf{Number of Images/} \\ \textbf{Number of Subjects}\\ \textbf{(Test Set)}\end{tabular}} \\ \hline \hline
\multirow{2}{*}{Apple iPhone 5}                                                     & \multirow{2}{*}{\begin{tabular}[c]{@{}c@{}}Device 1 \\ UNIT I\end{tabular}}  & Front (F) & 960$\times$1280   & 55/7                                                                                                                 & 344/41                                                                                                           \\
                                                                              &                                                                              & Rear (R)  & 1536$\times$2048  & 55/7                                                                                                                 & 355/41                                                                                                           \\
\multirow{2}{*}{Apple iPhone 5}                                                     & \multirow{2}{*}{\begin{tabular}[c]{@{}c@{}}Device 1 \\ UNIT II\end{tabular}} & Front (F) & 960$\times$1280   & 55/6                                                                                                                 & 164/20                                                                                                           \\
                                                                              &                                                                              & Rear (R)  & 2448$\times$3264  & 55/6                                                                                                                 & 170/20                                                                                                           \\
\multirow{2}{*}{\begin{tabular}[c]{@{}c@{}}Samsung Galaxy\\  S4\end{tabular}} & \multirow{2}{*}{\begin{tabular}[c]{@{}c@{}}Device 2 \\ UNIT I\end{tabular}}  & Front (F) & 1080$\times$1920  & 55/5                                                                                                                 & 577/69                                                                                                           \\
                                                                              &                                                                              & Rear (R)  & 2322$\times$4128  & 55/5                                                                                                                 & 600/70                                                                                                           \\ \hline 
                                                                              
                                                                             Samsung Galaxy S6 Edge & \textemdash & Front (F)   & 1080$\times$1920 & 55/6 &0/0 \\                                                                        
                                                                               HTC Desire EYE & \textemdash & Front (F)   & 1080$\times$1920 & 55/6 &0/0 \\                                                                        
                                                                                MEIZU X5 & \textemdash & Front (F)   & 1080$\times$1920 & 55/6 &0/0 \\                                                                        
                                                                                ASUS Zenfone Selfie &\textemdash & Front (F)   & 1080$\times$1920 & 55/6 &0/0 \\                                                                                                                                                       Sony XPERIA C5 Ultra Dual & \textemdash & Front (F)   & 1080$\times$1920 & 55/6 &0/0 \\                                                                        
                                                                                   Oppo N3 & \textemdash & Front (F)   & 1080$\times$1920 & 55/6 &0/0 \\                                                                        \hline \hline
                                                                              
\multicolumn{4}{|l||}{\textbf{TOTAL}}                                                                                                                                                             & 660/72                                                                                                               & 2,210/261 \\ \hline                                                                                                      
\end{tabular}}
\end{table*}
\noindent {\bf PRNU Spoofing:} We want the sensor classifier to assign an image belonging to source sensor $S$ to a specific target sensor $T$. To accomplish this task, we perform the following steps.\\
\noindent (i) First, we compute the parameter $\alpha$. Next, we transform the original image $\bm{I}$ from the source sensor to the DCT domain and then extract its low frequency components (as done in Algorithm~\ref{alg:Anon}). \\
(ii) A set of $N$ candidate images, $\bm{G}^i$, $i=[1,\cdots,N]$, belonging to the target sensor is selected, and each of them is subjected to DCT resulting in $\bm{G}_{dct}^i$ (see Section~\ref{Expts}). Next, we extract the high frequency coefficients from each $\bm{G}_{dct}^i$, apply inverse DCT, and then compute their average to yield $\bm{T}_{high}$. This averaged output represents the sensor traces of the target sensor. \\
(iii) Finally, we insert the averaged high frequency coefficients into $\bm{I}$ to generate $\bm{I'}$ which will now be classified as belonging to the target sensor, resulting in PRNU spoofing. \\
\tab The implementation details for PRNU spoofing are described in Algorithm~\ref{alg:Spoof}. The process of PRNU spoofing is illustrated using an example image in Figure~\ref{Fig: PRNUSpoof}.

\section{Experiments and Results}
\label{sec:expts}

\subsection{Dataset}
\label{Dataset}

\begin{figure}[h]
\centering
\subfloat[]
{   
    \includegraphics[scale=0.76]{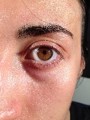} 
}
\hfill
\subfloat[]
{ 
    \includegraphics[scale=0.76]{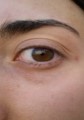}
}
 \hfill
\subfloat[]
{ 
    \includegraphics[scale=0.78]{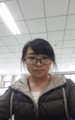} 
}
\hfill
\subfloat[]
{ 
    \includegraphics[scale=0.78]{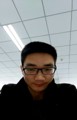} 
} \\
\subfloat[]
{ 
    \includegraphics[scale=0.78]{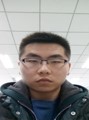} 
}
\hfill
\subfloat[]
{ 
    \includegraphics[scale=0.78]{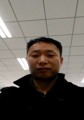} 
}
\hfill
\subfloat[]
{ 
    \includegraphics[scale=0.79]{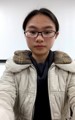} 
}
\hfill
\subfloat[]
{ 
    \includegraphics[scale=0.79]{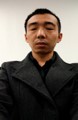} 
}
\caption{Example images from the MICHE-I and the OULU-NPU datasets acquired using (a) Apple iPhone 5 Rear, (b) Samsung Galaxy S4 Front, (c) Samsung Galaxy S6 Edge Front, (d) HTC Desire EYE Front, (e) MEIZU X5 Front, (f) ASUS Zenfone Selfie Front, (g) Sony XPERIA C5 Ultra Dual Front and (h) OPPO N3 Front sensors.}
\label{Fig: Example_images}
\end{figure}

We used the Mobile Iris Challenge Evaluation (MICHE-I) dataset~\cite{MICHE} and the OULU-NPU face dataset~\cite{Oulu_1, Oulu_2} for performing the experiments in this work. The MICHE-I dataset comprises of over 3,000 eye images from three devices: Apple iPhone 5, Samsung Galaxy S4 and Samsung Galaxy Tab 2~\cite{Galdi_PRL_15}. However, in our work, we employed the periocular images from two smartphones, Apple iPhone 5 and Samsung Galaxy S4, only. The authors in~\cite{Galdi_PRL_15} discovered that two separate units of Apple iPhone 5 were used for data collection. We refer to them as Unit I and Unit II respectively. Further, the images in the dataset were acquired using the front and rear camera sensors, separately. {\bf Thus, the MICHE-I dataset used in this work consists of data from 6 sensors}. The OULU-NPU face dataset comprises of 4,950 face videos recorded using the front cameras of six mobile devices\textemdash Samsung Galaxy S6 Edge, HTC Desire EYE, MEIZU X5, ASUS Zenfone Selfie, Sony XPERIA C5 Ultra Dual and OPPO N3. The videos were recorded in three sessions with different illumination and background scenes. {\bf We only use the bonafide face videos/images in the OULU-NPU dataset corresponding to 6 sensors.} See Figure~\ref{Fig: Example_images}. The specifications of the dataset are described in Table~\ref{Tab: Dataset}. 

We split each dataset into a training set and a test set. We followed a subject-disjoint protocol for creating the training and test sets. The images in the training set are used for generating the reference pattern for each sensor, as indicated in the fifth column of Table~\ref{Tab: Dataset}. Our training set consists of 55 images~\cite{Ross_18} from each camera sensor in the MICHE-I dataset. The OULU-NPU database contains videos, and so we selected 55 frames (20 frames from the first session, 20 frames from the second, and 15 frames from the third) from 6 subjects, for each of the 6 sensors. The test set comprises of images belonging to the MICHE-I dataset only (see the last column in Table~\ref{Tab: Dataset}). 
Thus, our dataset consists of 2,870 images corresponding to 333 subjects acquired using 12 camera sensors. Next, we describe the experiments conducted in this work. 

\subsection{Experimental Methodology}
\label{Expts}

For the \textit{sensor de-identification} experiments, we first computed the sensor reference patterns from the traning set for each of the 12 sensors (see Table~\ref{Tab: Dataset}) using the three PRNU estimation schemes \textit{viz.}, Enhanced PRNU, MLE PRNU and Phase PRNU. Next, we used a small number of images (=10) as the validation set to compute the parameter $\eta =[0,1]$ to be used for PRNU anonymization and PRNU spoofing, separately. We estimated $\eta=0.9$ for PRNU anonymization and $\eta=0.7$ for PRNU spoofing. The test experiments were conducted on images belonging to the MICHE-I dataset only. \textbf{However, the evaluation process involved all the 12 sensor reference patterns.} The experiments evaluated three PRNU estimation schemes: Enhanced PRNU,\footnote{We employed Enhancement Model III and we set the user defined threshold to 6~\cite{Li_TIFS_10, Ross_18}.} MLE PRNU and Phase PRNU methods. We used normalized cross-correlation for sensor identification. For the PRNU spoofing experiments, the source and target sensors were from the MICHE-I dataset and were either both front or both rear sensors. Thus, there were $2 \times \Myperm[3]{2} = 12$ PRNU spoofing experiments. Due to the significant difference in resolutions between the front and rear sensors of smartphones, we did not perform front-to-rear or rear-to-front spoofing. We selected $N$, \ie, the number of candidate images belonging to the target sensor (see Algorithm~\ref{alg:Spoof}), to be the number of test images for that sensor (see the last column in Table~\ref{Tab: Dataset}).

\begin{table*}[h]
\centering
\caption{Performance of the proposed algorithm for \textbf{PRNU Anonymization} in terms of sensor identification accuracy (\%). Results are evaluated using 3 PRNU estimation schemes. `Original' corresponds to sensor identification using images prior to perturbation. `After' corresponds to sensor identification using images after perturbation and `Change' indicates the difference between the `Original' and `After' sensor identification accuracies. A high positive value in the `Change' field indicates successful PRNU Anonymization. }
\label{Tab:PRNU Anonymization}
\scalebox{0.82}{
\begin{tabular}{|l|l|lll|lll|lll|}
\hline
\multicolumn{1}{|c|}{\multirow{2}{*}{\textbf{Device Identifier}}} & \multirow{2}{*}{\textbf{Sensors}} & \multicolumn{3}{c|}{\textbf{Enhanced PRNU}} & \multicolumn{3}{c|}{\textbf{MLE PRNU}} & \multicolumn{3}{c|}{\textbf{Phase PRNU}} \\ 
\multicolumn{1}{|c|}{}                                             &                                   & Original       & After      & Change         & Original    & After     & Change       & Original      & After      & Change       \\ \hline
\multirow{2}{*}{Device 1 UNIT I}                                 & Front                             & 99.71        &  18.31    &  81.40      &  99.71     &  17.73  &    81.98     & 99.71      &  22.67     &  77.04       \\
                                                                 & Rear                              & 99.51 &     16.06 &  83.45       & 97.32      & 16.06    & 81.26       & 98.05       & 21.69     & 76.36        \\ \hline
\multirow{2}{*}{Device 1 UNIT II}                                & Front                             & 96.34        &   21.34    &   75        & 96.34      &  25.61    &  70.73      & 93.90      &   26.83    &  67.07       \\
                                                                 & Rear                              & 94.71 &   11.76   &  82.95         & 88.24      & 14.12     & 74.12       & 87.65       & 11.76     &  75.89       \\ \hline
\multirow{2}{*}{Device 2 UNIT I}                                 & Front                             & 100          &  3.81    &  96.19         & 100        & 5.72    &   94.28     & 100         & 13.69      &  86.31       \\
                                                                 & Rear                              & 100        &   4.50  &   95.50        & 100      &  3.17   &    96.83       & 100       &  6.50     &   93.50      \\ \hline \hline
\multicolumn{2}{|l|}{\textbf{AVERAGE}}                                                                 &              &            & 85.75      &            &           &  83.20    &             &            & 79.36 \\ \hline   
\end{tabular}}
\end{table*}

\begin{table*}[t]
\centering
\caption{Performance of the proposed algorithm for \textbf{PRNU Spoofing} in terms of spoof success rate (SSR). Results are evaluated using 3 PRNU estimation schemes. A high SSR value indicates successful spoofing.}
\label{Tab:PRNU Spoofing}
\scalebox{0.82}{
\begin{tabular}{|l|l|lll|} \hline
\multirow{2}{*}{\textbf{Source Sensor}}                                                    & \multirow{2}{*}{\textbf{Target Sensor}}                                   & \multicolumn{3}{c|}{\textbf{Spoof Success Rate (\%)}} \\
                                                                                  &                                                                  & \textbf{Enhanced PRNU }  & \textbf{MLE PRNU}   & \textbf{Phase PRNU } \\ \hline 
\multirow{2}{*}{\begin{tabular}[c]{@{}l@{}}Device 1 UNIT I\\ FRONT\end{tabular}}  & Device 1 UNIT II FRONT & 100          &  100     &  100        \\
                                                                                  & Device 2 UNIT I FRONT  & 96.80 &  100      &  100      \\ \hline
\multirow{2}{*}{\begin{tabular}[c]{@{}l@{}}Device 1 UNIT II\\ FRONT\end{tabular}} & Device 1 UNIT I FRONT  & 100             & 100       &   100       \\
                                                                                  & Device 2 UNIT I FRONT & 97.56 & 100       &  100       \\ \hline
\multirow{2}{*}{\begin{tabular}[c]{@{}l@{}}Device 2 UNIT I\\ FRONT\end{tabular}}  & Device 1 UNIT I FRONT & 100             &   100     & 100         \\
                                                                                  & Device 1 UNIT II FRONT &100  &100  &100          \\ \hline
\multirow{2}{*}{\begin{tabular}[c]{@{}l@{}}Device 1 UNIT I\\ REAR\end{tabular}}   & Device 1 UNIT II REAR  &  100           & 100       &    94.08    \\
                                                                                  & Device 2 UNIT I REAR   & 99.44 &   96.90   &  97.46     \\ \hline
\multirow{2}{*}{\begin{tabular}[c]{@{}l@{}}Device 1 UNIT II\\ REAR\end{tabular}}  & Device 1 UNIT I REAR   &  100        &  100       &   100       \\
                                                                                  & Device 2 UNIT I REAR   & 100 &   98.82      &  100       \\ \hline
\multirow{2}{*}{\begin{tabular}[c]{@{}l@{}}Device 2 UNIT I\\ REAR\end{tabular}}   & Device 1 UNIT I REAR  & 100            & 100        &   100       \\
                                                                                  & Device 1 UNIT II REAR &100  &        100 &  99.83 \\ \hline \hline
                                                                                  
\multicolumn{2}{|l|}{\textbf{AVERAGE}}                                                                              &   99.48                &  99.64          & 99.28 \\ \hline   
\end{tabular}}
\end{table*}

\begin{figure*}
\subfloat[]{ 
\hspace{0.4cm}\includegraphics[scale=.14]{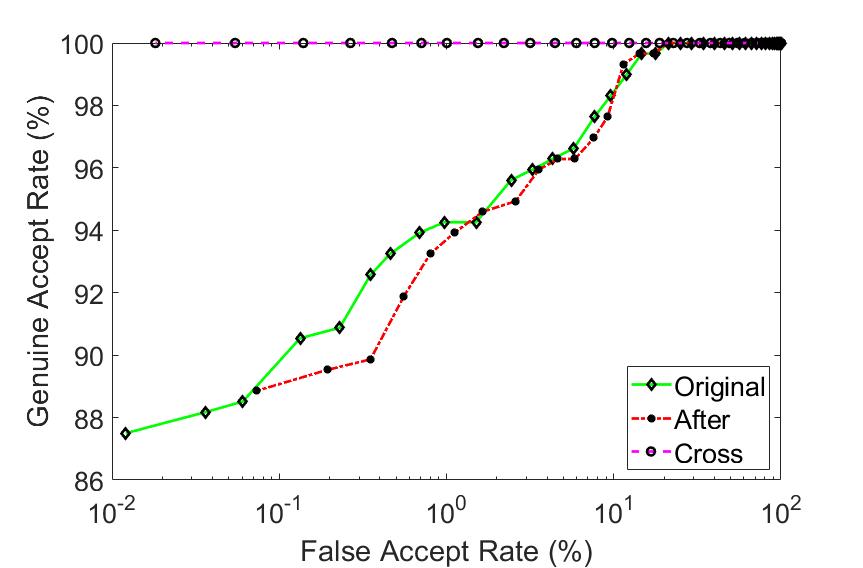} \hspace{-2.9cm}\raisebox{\dimexpr 2.6cm+\height}{\footnotesize{Front-Indoor}} \hspace{1cm} 
\includegraphics[scale=.14]{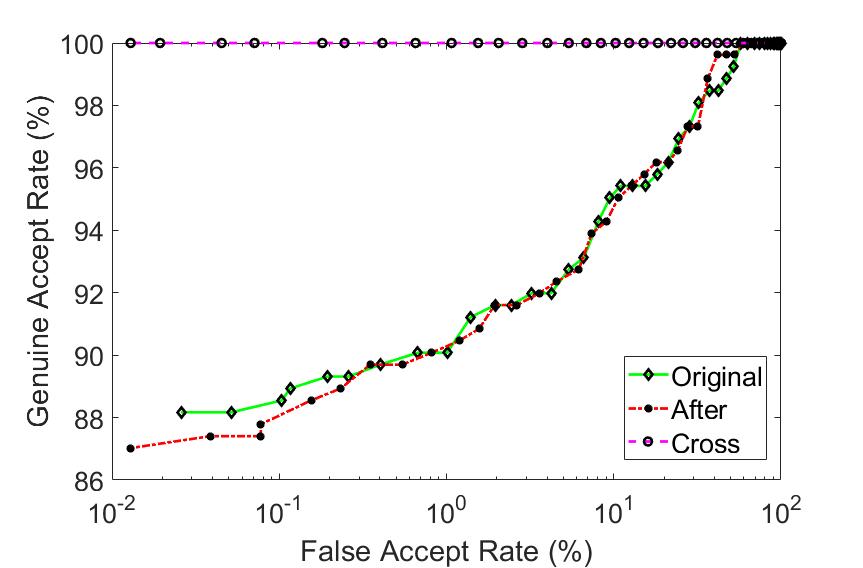} \hspace{-2.9cm}\raisebox{\dimexpr 2.6cm+\height}{\footnotesize{Front-Outdoor}} \hspace{1cm}  \\ 
\includegraphics[scale=.14]{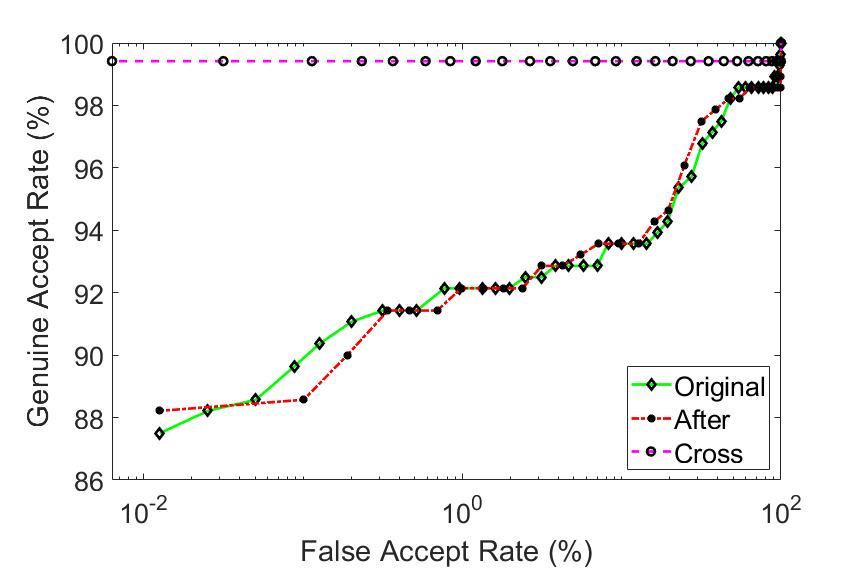} \hspace{-2.9cm}\raisebox{\dimexpr 2.6cm+\height}{\footnotesize{Rear-Indoor}} \hspace{1cm}
\includegraphics[scale=.14]{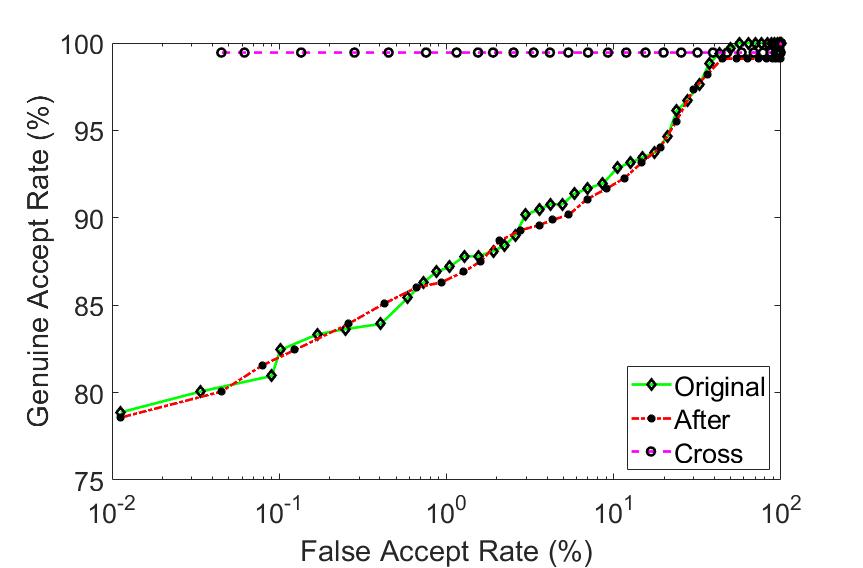} \hspace{-2.9cm}\raisebox{\dimexpr 2.6cm+\height}{\footnotesize{Rear-Outdoor}} \hspace{1cm} 
}
\vspace{-0.1cm}
\subfloat[]{ 
\hspace{0.4cm}\includegraphics[scale=.14]{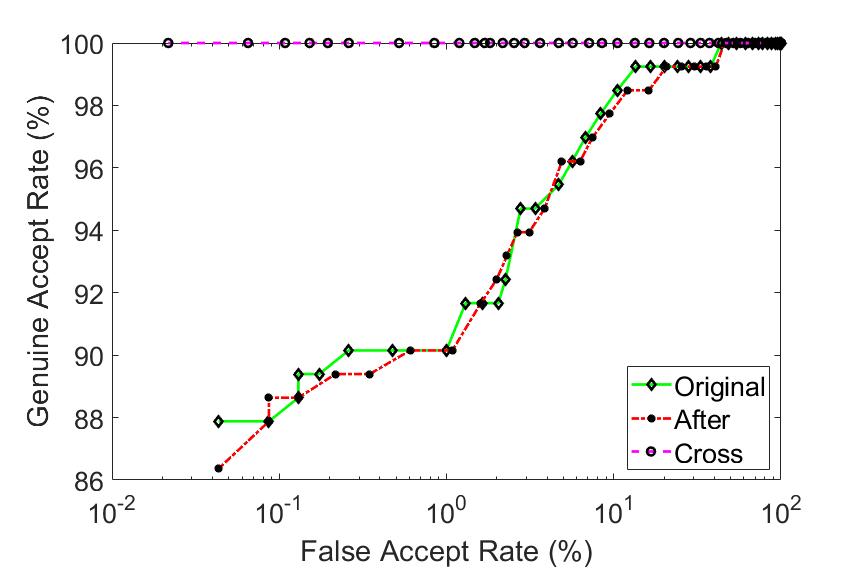} \hspace{-2.9cm}\raisebox{\dimexpr 2.6cm+\height}{\footnotesize{Front-Indoor}} \hspace{1cm} 
\includegraphics[scale=.14]{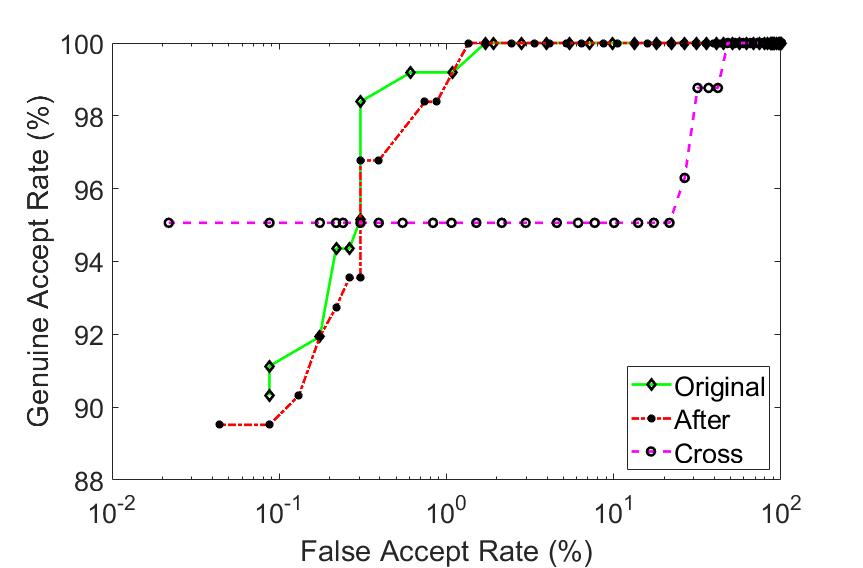} \hspace{-2.9cm}\raisebox{\dimexpr 2.6cm+\height}{\footnotesize{Front-Outdoor}} \hspace{1cm}  \\
\includegraphics[scale=.14]{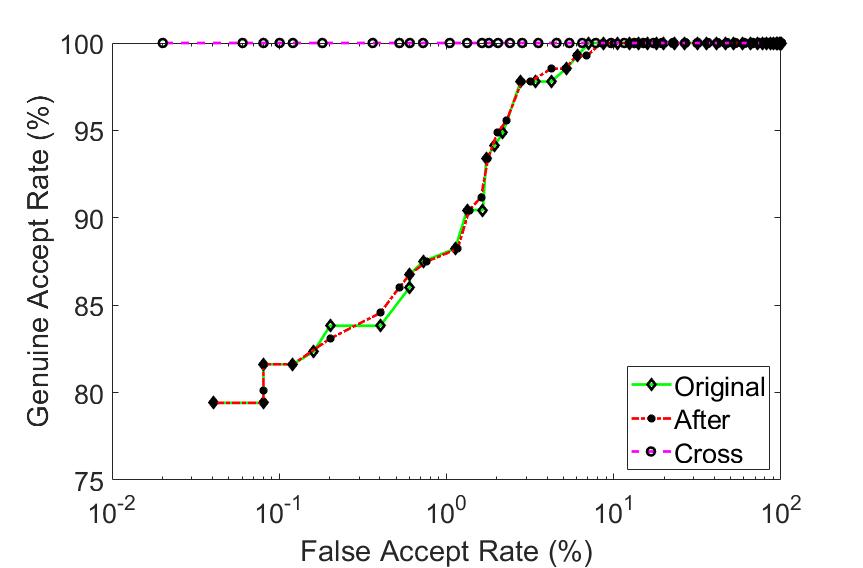} \hspace{-2.9cm}\raisebox{\dimexpr 2.6cm+\height}{\footnotesize{Rear-Indoor}} \hspace{1cm}
\includegraphics[scale=.14]{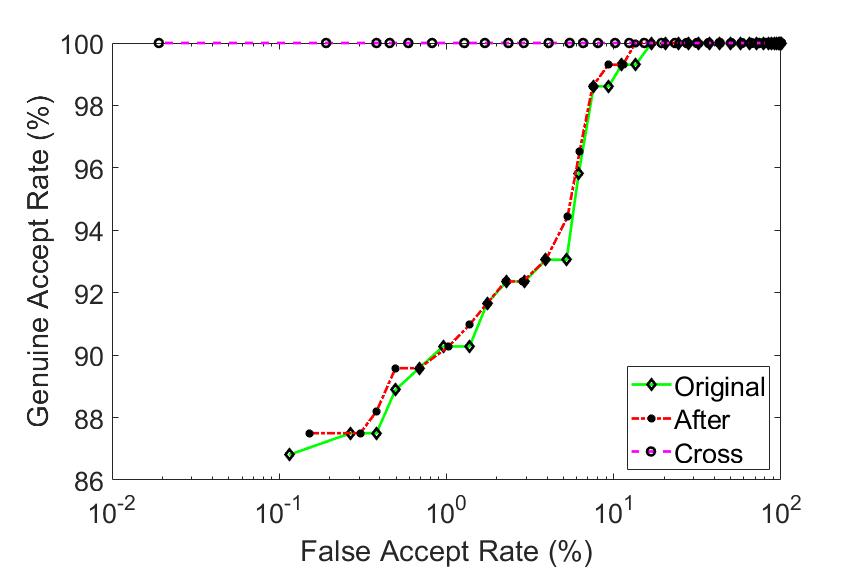} \hspace{-2.9cm}\raisebox{\dimexpr 2.6cm+\height}{\footnotesize{Rear-Outdoor}} \hspace{1cm} 
}
\vspace{-0.1cm}
\subfloat[]{ 
\hspace{0.4cm}\includegraphics[scale=.14]{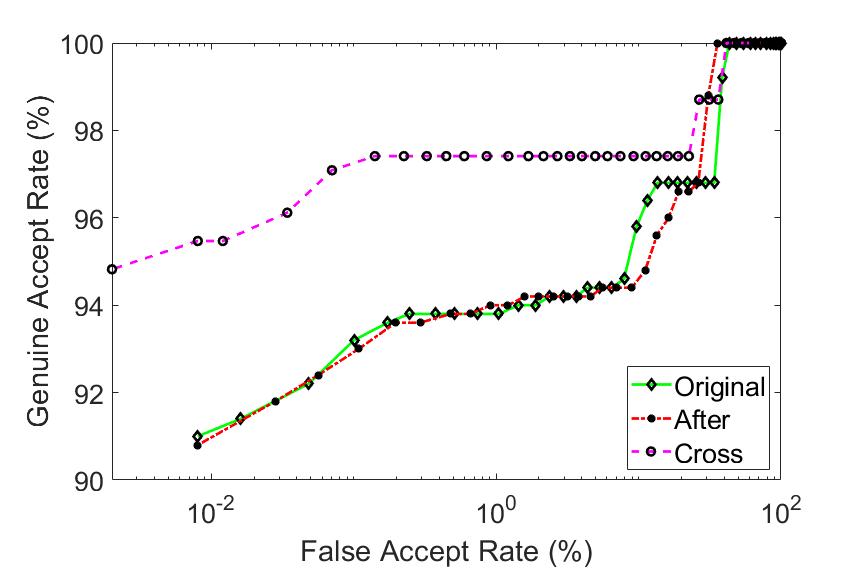}  \hspace{-2.9cm}\raisebox{\dimexpr 2.6cm+\height}{\footnotesize{Front-Indoor}} \hspace{1cm} 
\includegraphics[scale=.14]{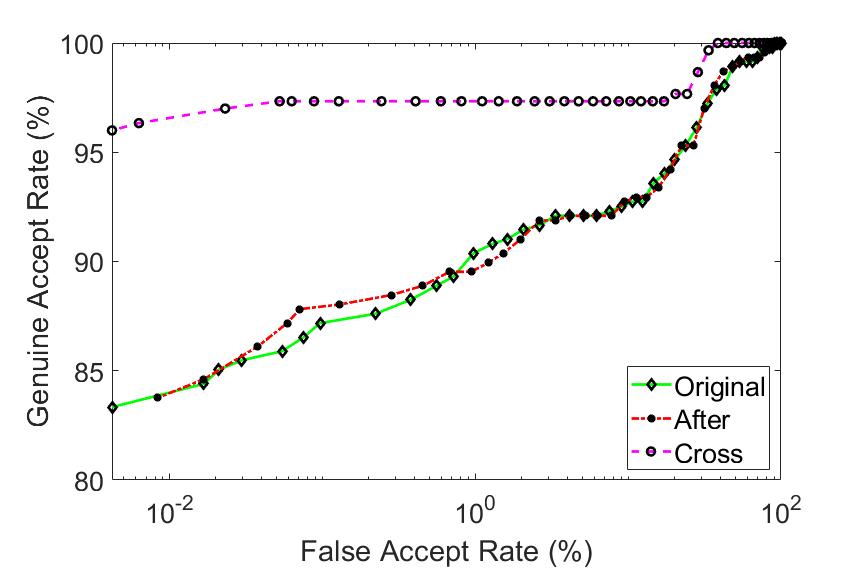}  \hspace{-2.9cm}\raisebox{\dimexpr 2.6cm+\height}{\footnotesize{Front-Outdoor}} \hspace{1cm}  \\
\includegraphics[scale=.14]{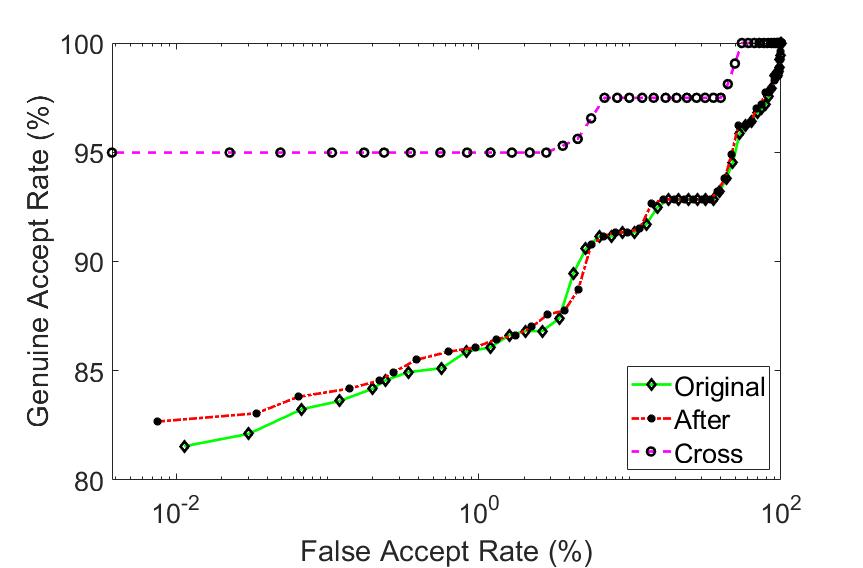}  \hspace{-2.9cm}\raisebox{\dimexpr 2.6cm+\height}{\footnotesize{Rear-Indoor}} \hspace{1cm} 
\includegraphics[scale=.14]{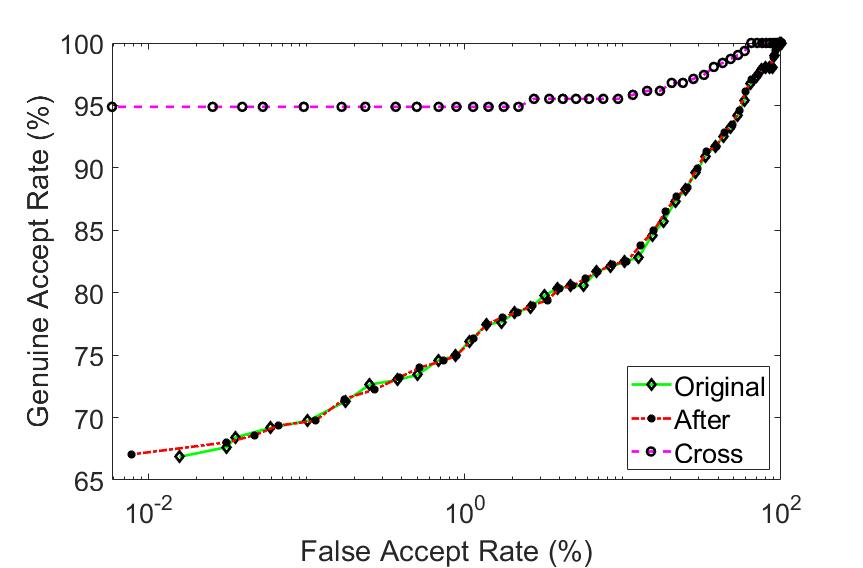}  \hspace{-2.9cm}\raisebox{\dimexpr 2.6cm+\height}{\footnotesize{Rear-Outdoor}} \hspace{1cm} 
}

\caption{ROC curves for matching \textbf{PRNU Anonymized} images. Each row corresponds to a different device identifier: (a) Device 1 UNIT I, (b) Device 1 UNIT II and (c) Device 2 UNIT I. }
\label{Fig: ROC_PRNU Anon}
\end{figure*}

\begin{figure*}[]
\subfloat[]{
\hspace{1.0cm} \includegraphics[scale=.13]{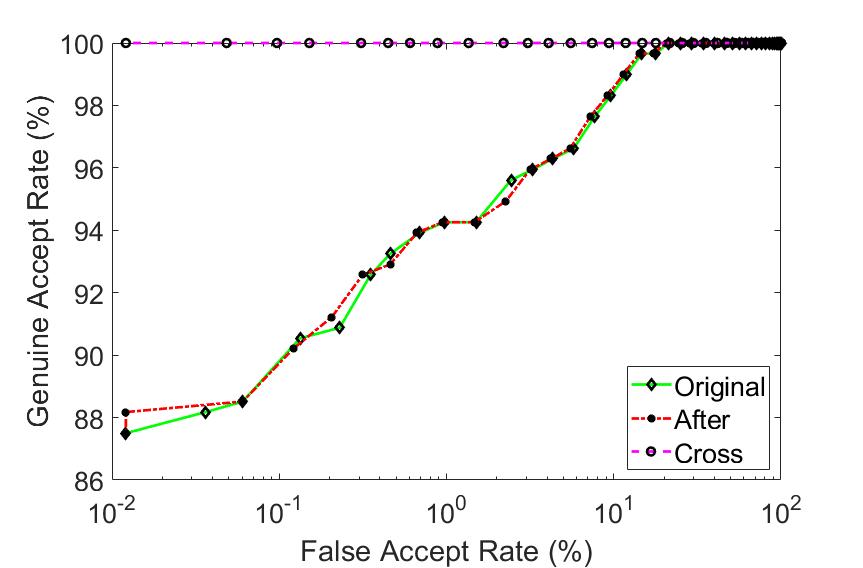}  \hspace{-2.9cm}\raisebox{\dimexpr 2.6cm+\height}{\footnotesize{Front-Indoor}} \hspace{1cm} 
\includegraphics[scale=.13]{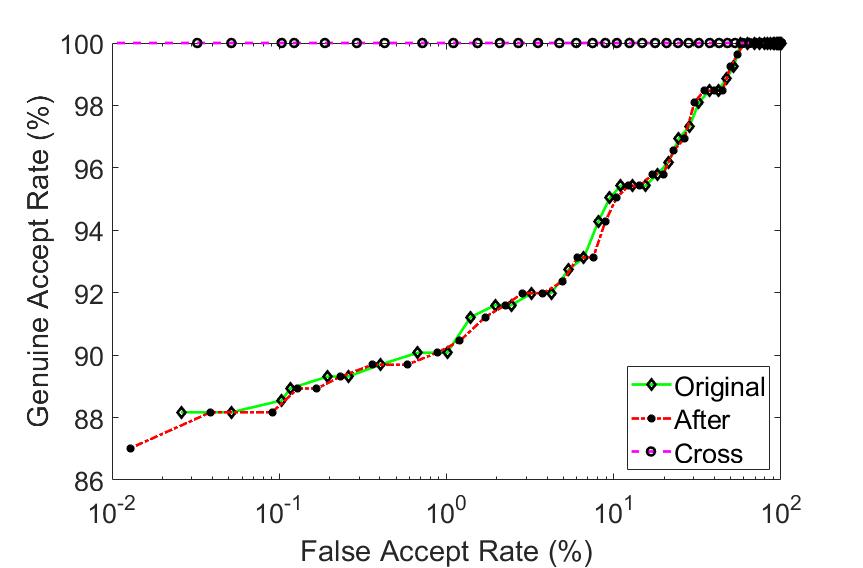}  \hspace{-2.9cm}\raisebox{\dimexpr 2.6cm+\height}{\footnotesize{Front-Outdoor}} \hspace{1cm}   \\ 
\includegraphics[scale=.13]{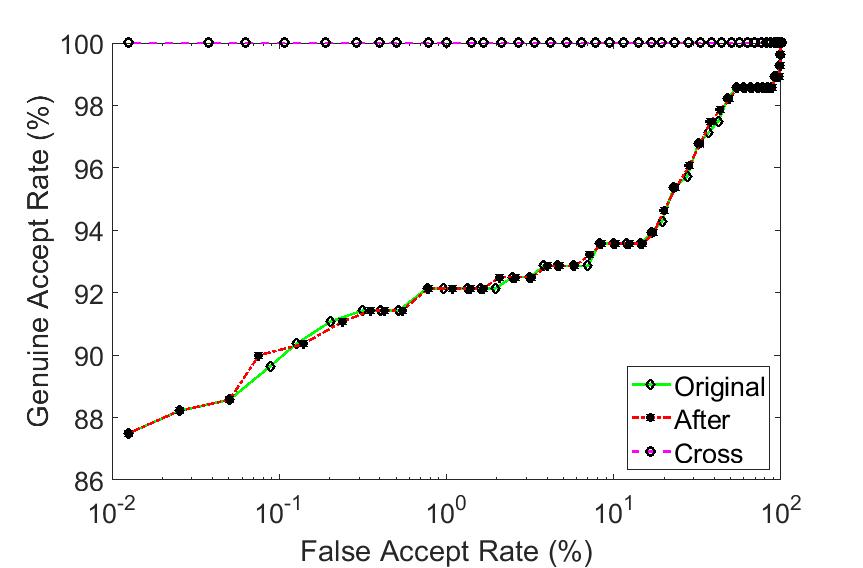}  \hspace{-2.9cm}\raisebox{\dimexpr 2.6cm+\height}{\footnotesize{Rear-Indoor}} \hspace{1cm} 
\includegraphics[scale=.13]{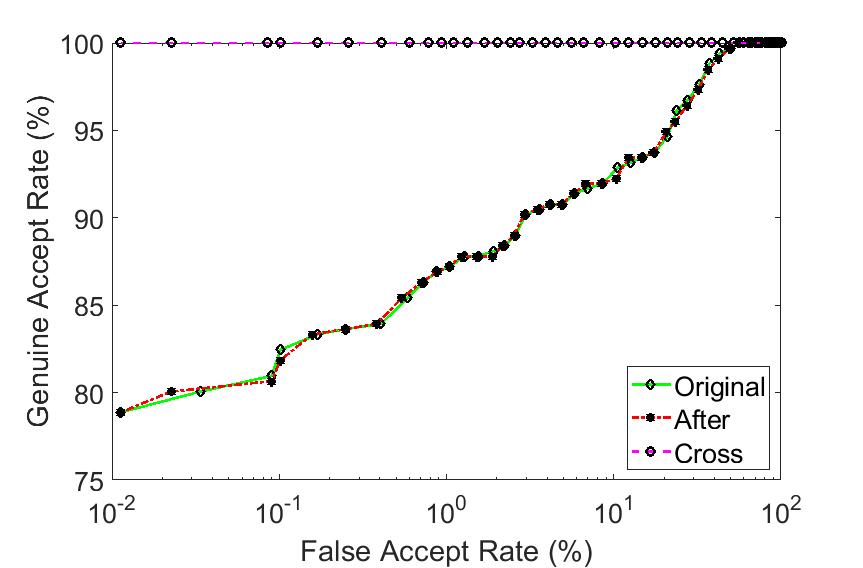} \hspace{-2.9cm}\raisebox{\dimexpr 2.6cm+\height}{\footnotesize{Rear-Outdoor}}  \hspace{1cm}
}
\vspace{-0.1cm}
\subfloat[]{ 
\hspace{1.0cm} \includegraphics[scale=.13]{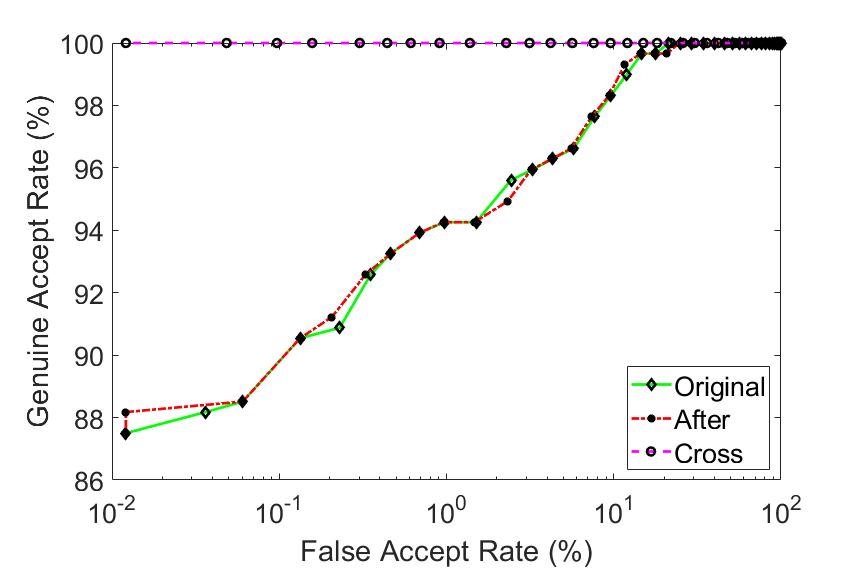}  \hspace{-2.9cm}\raisebox{\dimexpr 2.6cm+\height}{\footnotesize{Front-Indoor}} \hspace{1cm}  
\includegraphics[scale=.13]{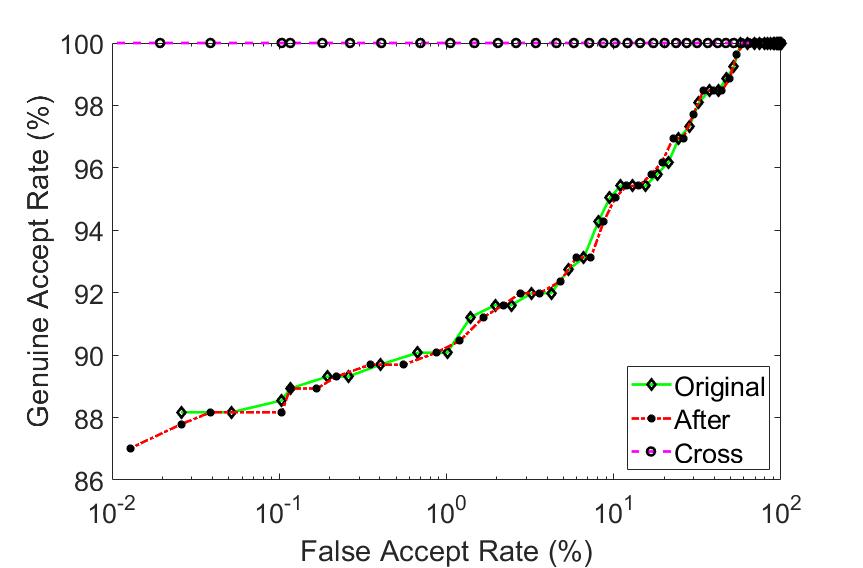} \hspace{-2.9cm}\raisebox{\dimexpr 2.6cm+\height}{\footnotesize{Front-Outdoor}} \hspace{1cm} \\
\includegraphics[scale=.13]{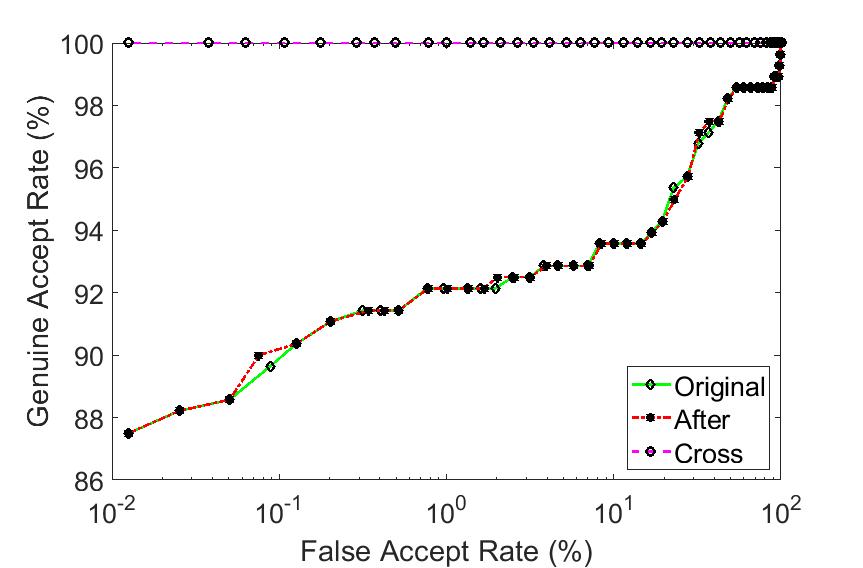}  \hspace{-2.9cm}\raisebox{\dimexpr 2.6cm+\height}{\footnotesize{Rear-Indoor}} \hspace{1cm}
\includegraphics[scale=.13]{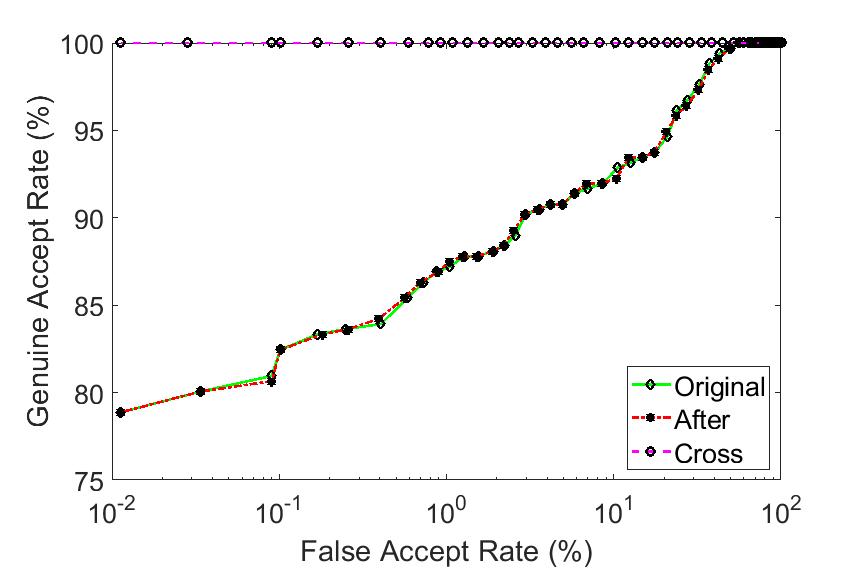} \hspace{-2.9cm}\raisebox{\dimexpr 2.6cm+\height}{\footnotesize{Rear-Outdoor}} \hspace{1cm}
}

\caption{ROC curves for matching \textbf{PRNU Spoofed} images. Here, the source sensor is Device 1 UNIT I. In this case, the target sensors are: (a) Device 1 UNIT II (top row) and (b) Device 2 UNIT I (bottom row). }
\label{Fig: ROC_PRNUspoof1}
\end{figure*}

For the \textit{biometric matching} experiments, we considered a periocular matcher, as many of the images used in this work are partial face images. We employed the ResNet-101~\cite{ResNet101} architecture pre-trained on ImageNet~\cite{imageNet} dataset for performing periocular matching. We utilized the features from layer 170 which were shown to perform the best for periocular matching in~\cite{ResNet}. We applied Contrast Limited Adaptive Histogram Equalization (CLAHE) to the images before feeding them to the convolutional neural network. We used the cosine similarity for computing the match score between the probe and gallery images. We performed three sets of matching experiments, \textit{viz.}, (i) original: both probe and gallery images comprise of unmodified images, (ii) after: both probe and gallery images comprise of modified images and (iii) cross: the gallery images are the original samples while the probe images are the modified images and the genuine scores are computed by utilizing 2 sample images (belonging to the same subject), \ie, the original image and the modified image; the impostor scores are computed by taking pairs of samples belonging to different subjects. Furthermore, we conducted experiments separately for the two acquisition settings in this database: Indoor and Outdoor. 

\subsection{Results}
\label{res}

For the sensor de-identification experiments, we used sensor identification accuracy as the evaluation metric for PRNU anonymization and the spoof success rate (SSR) as the evaluation metric for the PRNU spoofing algorithm. For PRNU anonymization, we first compute the sensor identification accuracy of the original images. Before perturbation, the images are assigned to the correct sensor with high accuracies by all three PRNU estimation schemes (see `Original' column in Table~\ref{Tab:PRNU Anonymization}). Next, when the sensor classifier accepts the modified images as input, the results indicate a significant degradation in the sensor identification accuracy for a majority of the cases (see `After' column in Table~\ref{Tab:PRNU Anonymization}). The differences in the sensor identification accuracies before and after perturbation are reported in the `Change' column in Table~\ref{Tab:PRNU Anonymization}. An average difference (change) of 82.77\% in the sensor identification accuracies between pre- and post-perturbed images is observed for all the three PRNU estimation schemes evaluated in this work (Enhanced PRNU: 85.75\%, MLE PRNU: 83.20\% and Phase PRNU: 79.36\%). The results indicate successful PRNU anonymization thereby ensuring sensor de-identification. 

The second set of results, pertaining to PRNU spoofing, reports the SSR for the perturbed images. SSR computes the proportion of perturbed images that are assigned to the target sensor. The results in Table~\ref{Tab:PRNU Spoofing} indicate successful spoofing with respect to all the PRNU estimation schemes considered in this work. An average SSR of 99.48\% is observed when evaluated using Enhanced PRNU, 99.64\% when evaluated using MLE PRNU, and 99.28\% when evaluated using Phase PRNU for all 12 PRNU spoofing experiments. The proposed spoofing experiment fails to confound the Phase PRNU estimation scheme, particularly when the source sensor is the rear sensor of Device 1 UNIT I and the target sensor is the rear sensor of Device 1 UNIT II. Upon analysis, we observed that the original images belonging to Device 1 UNIT II rear sensor resulted in the \textit{lowest} sensor identification accuracy for all three PRNU estimation schemes (see Table~\ref{Tab:PRNU Anonymization}). We speculate that the images may contain some artifacts that are interfering with reliable PRNU estimation as well as the spoofing process. Therefore, we performed another experiment where we increased the value of $\eta$ from 0.7 to 0.9 for that particular spoofing experiment and we observed that the SSR increased to 100\% for all 3 PRNU estimation schemes. However, visual analysis reveals that the spoofed images resulting from the two different values of $\eta$ have perceptible differences ($\eta$ = 0.9 results in a more blurred image than when $\eta$ = 0.7 is used). Finally, we studied the performance of our PRNU spoofing algorithm when a smaller number of candidate images, $N$, is employed (50\%, 10\% and 1\% of the test set). Surprisingly, even when only 1\% of the test set is used as candidate images, \ie, $N=4$, we observed an average SSR of 99.6\% across the three PRNU estimation schemes. However, the spoofed images are significantly degraded as they contain some spurious scene details from the candidate images (possibly, the averaging operation in Step 17 of Algorithm~\ref{alg:Spoof} suppresses scene details more aggressively for a high value of $N$).

\begin{figure*}[]
\subfloat[]{
\hspace{1.0cm} \includegraphics[scale=.13]{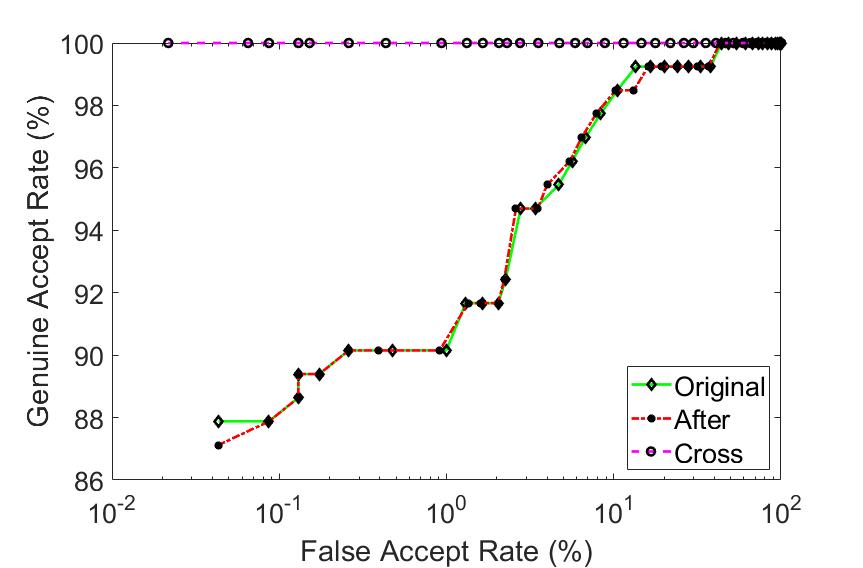}  \hspace{-2.9cm}\raisebox{\dimexpr 2.6cm+\height}{\footnotesize{Front-Indoor}} \hspace{1cm} 
\includegraphics[scale=.13]{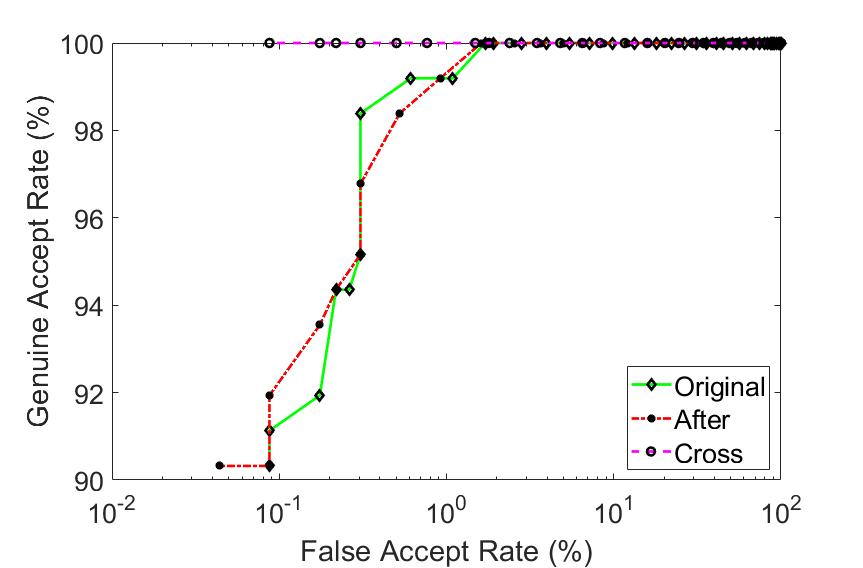}  \hspace{-2.9cm}\raisebox{\dimexpr 2.6cm+\height}{\footnotesize{Front-Outdoor}} \hspace{1cm}  \\ 
\includegraphics[scale=.13]{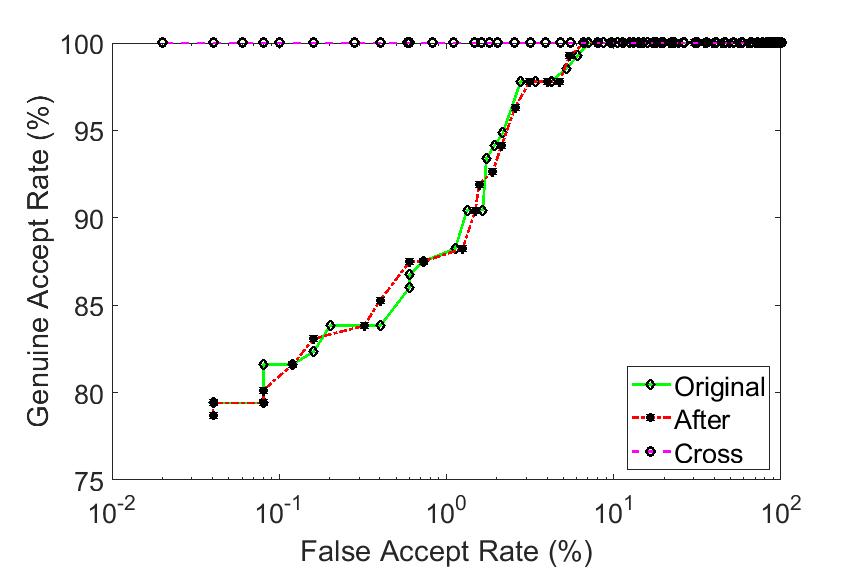}  \hspace{-2.9cm}\raisebox{\dimexpr 2.6cm+\height}{\footnotesize{Rear-Indoor}} \hspace{1cm}
\includegraphics[scale=.13]{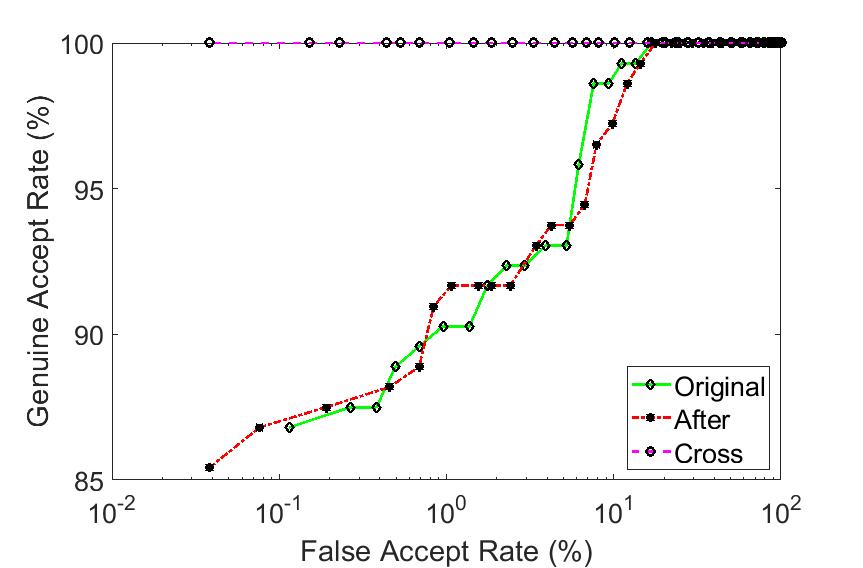}  \hspace{-2.9cm}\raisebox{\dimexpr 2.6cm+\height}{\footnotesize{Rear-Outdoor}} \hspace{1cm}
}
\vspace{-0.1cm}
\subfloat[]{ 
\hspace{1.0cm} \includegraphics[scale=.13]{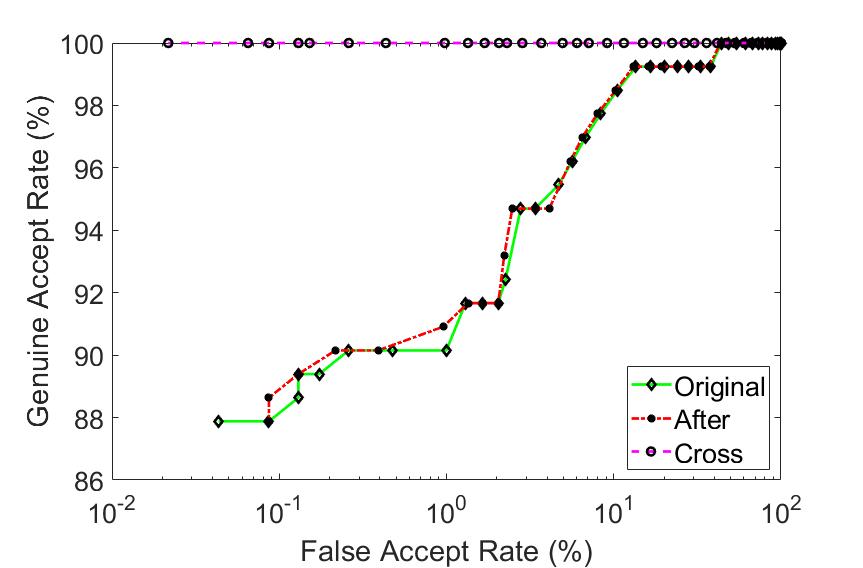}  \hspace{-2.9cm}\raisebox{\dimexpr 2.6cm+\height}{\footnotesize{Front-Indoor}} \hspace{1cm} 
\includegraphics[scale=.13]{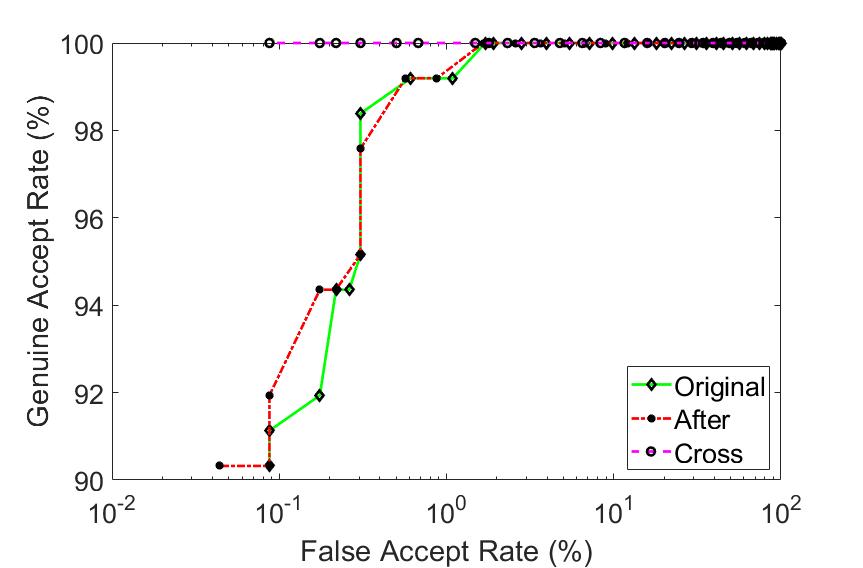} \hspace{-2.9cm}\raisebox{\dimexpr 2.6cm+\height}{\footnotesize{Front-Outdoor}} \hspace{1cm}      \\
\includegraphics[scale=.13]{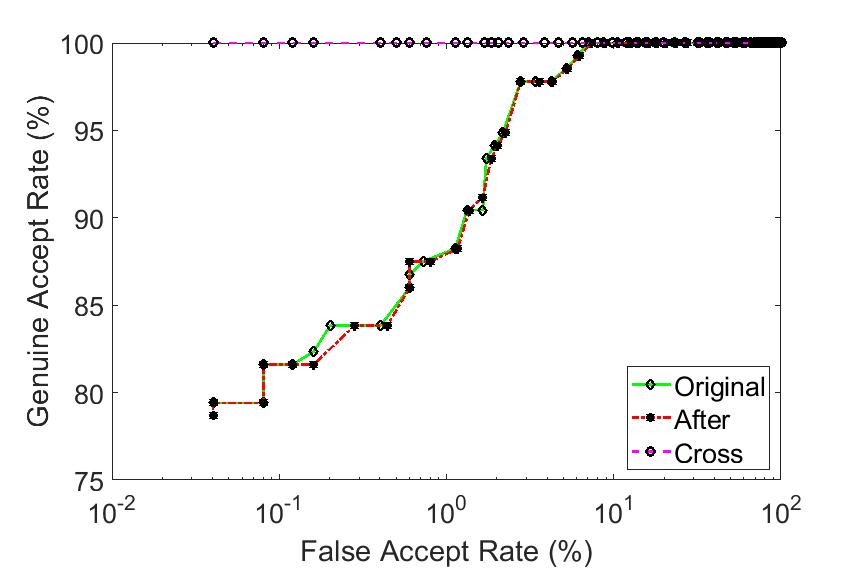}  \hspace{-2.9cm}\raisebox{\dimexpr 2.6cm+\height}{\footnotesize{Rear-Indoor}} \hspace{1cm}
\includegraphics[scale=.13]{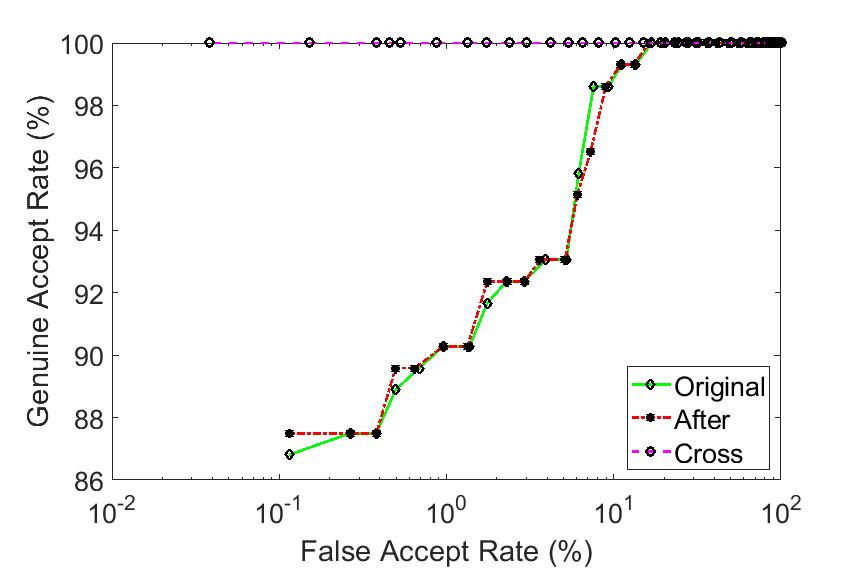}  \hspace{-2.9cm}\raisebox{\dimexpr 2.6cm+\height}{\footnotesize{Rear-Outdoor}} \hspace{1cm}
}

\caption{ROC curves for matching \textbf{PRNU Spoofed} images. Here, the source sensor is Device 1 UNIT II. In this case, the target sensors are: (a) Device 1 UNIT I (top row) and (b) Device 2 UNIT I (bottom row). }
\label{Fig: ROC_PRNUspoof2}
\end{figure*}

\begin{figure*}[]

\subfloat[]{
\hspace{1.0cm} \includegraphics[scale=.13]{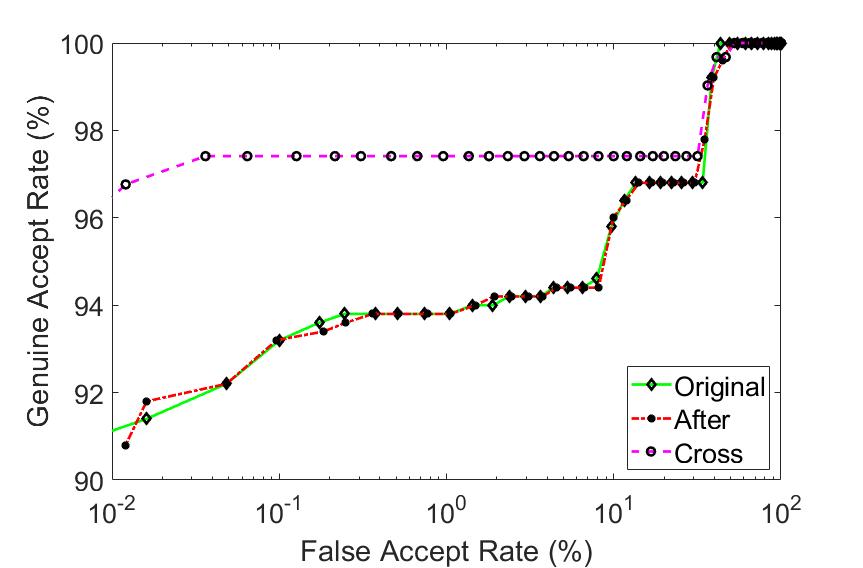}\hspace{-2.9cm}\raisebox{\dimexpr 2.6cm+\height}{\footnotesize{Front-Indoor}} \hspace{1cm} 
\includegraphics[scale=.13]{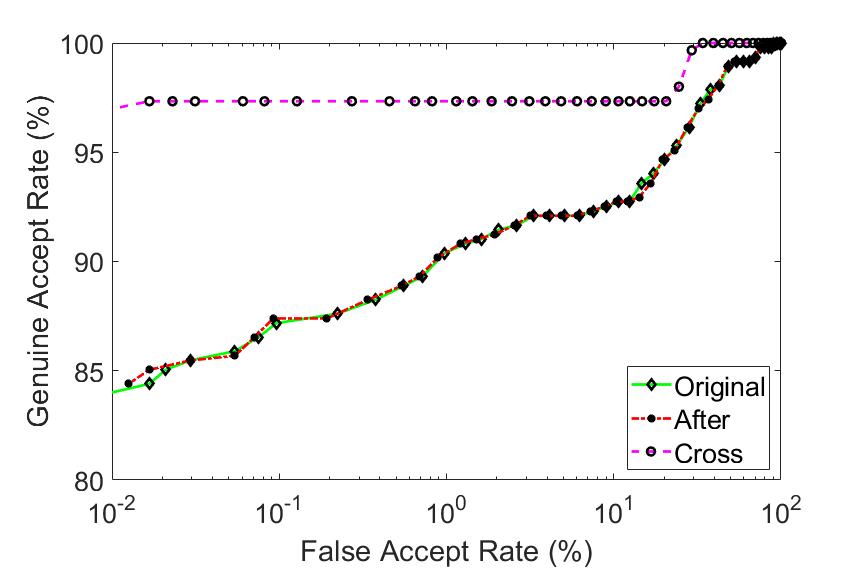} \hspace{-2.9cm}\raisebox{\dimexpr 2.5cm+\height}{\footnotesize{Front-Outdoor}} \hspace{1cm}  \\ 
\includegraphics[scale=.13]{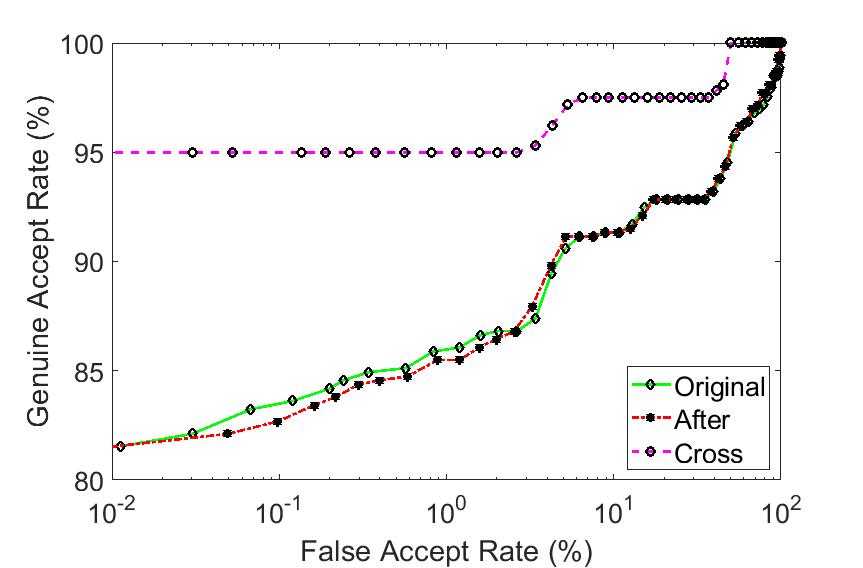} \hspace{-2.9cm}\raisebox{\dimexpr 2.6cm+\height}{\footnotesize{Rear-Indoor}} \hspace{1cm} 
\includegraphics[scale=.13]{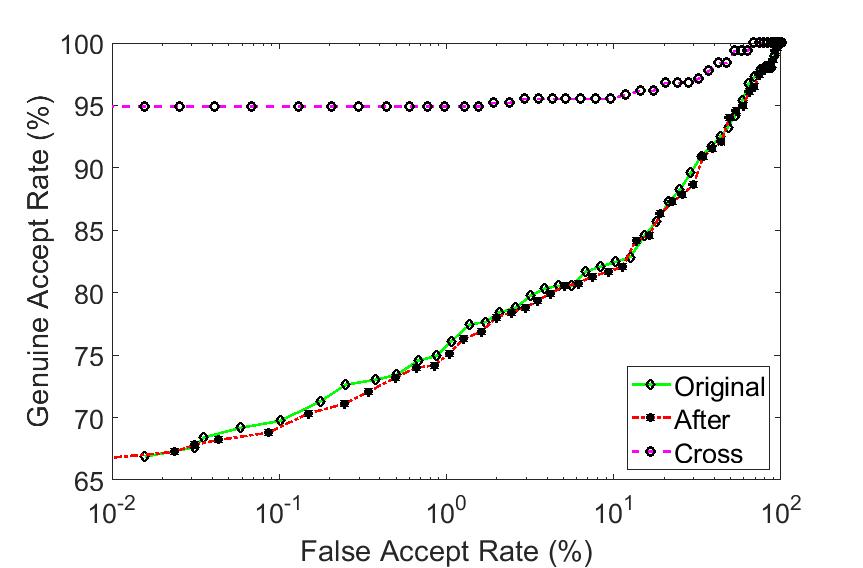} \hspace{-2.9cm}\raisebox{\dimexpr 2.6cm+\height}{\footnotesize{Rear-Outdoor}}  \hspace{1cm}
}
\vspace{-0.1cm}
\subfloat[]{
\hspace{1.0cm} \includegraphics[scale=.13]{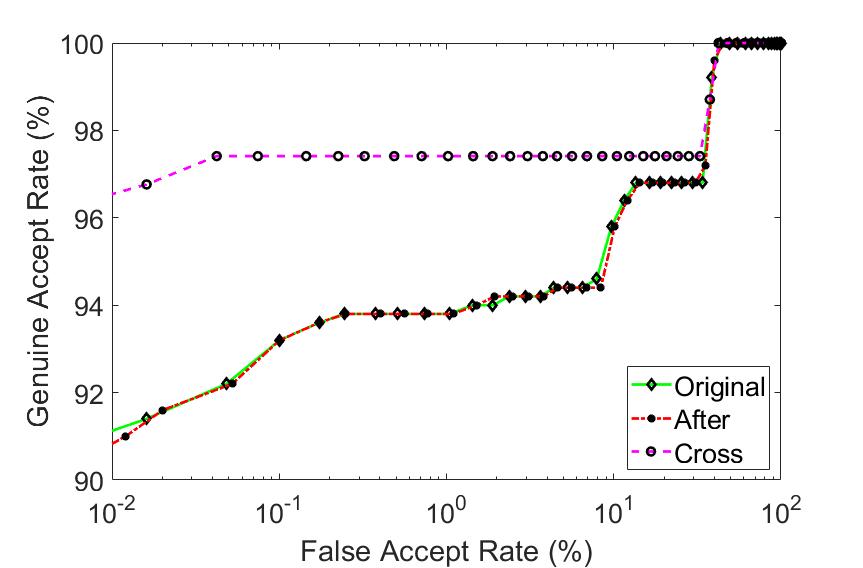} \hspace{-2.9cm}\raisebox{\dimexpr 2.6cm+\height}{\footnotesize{Front-Indoor}} \hspace{1cm} 
\includegraphics[scale=.13]{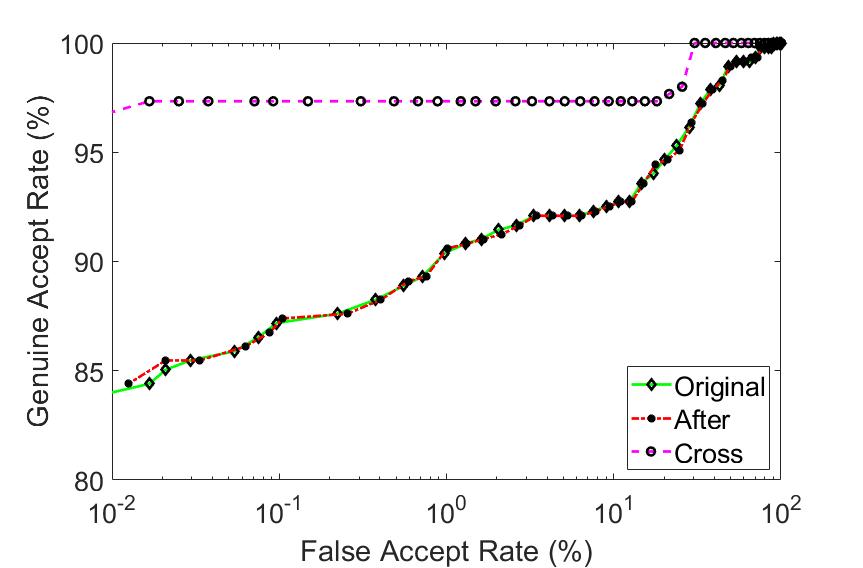} \hspace{-2.9cm}\raisebox{\dimexpr 2.5cm+\height}{\footnotesize{Front-Outdoor}} \hspace{1cm}  \\
\includegraphics[scale=.13]{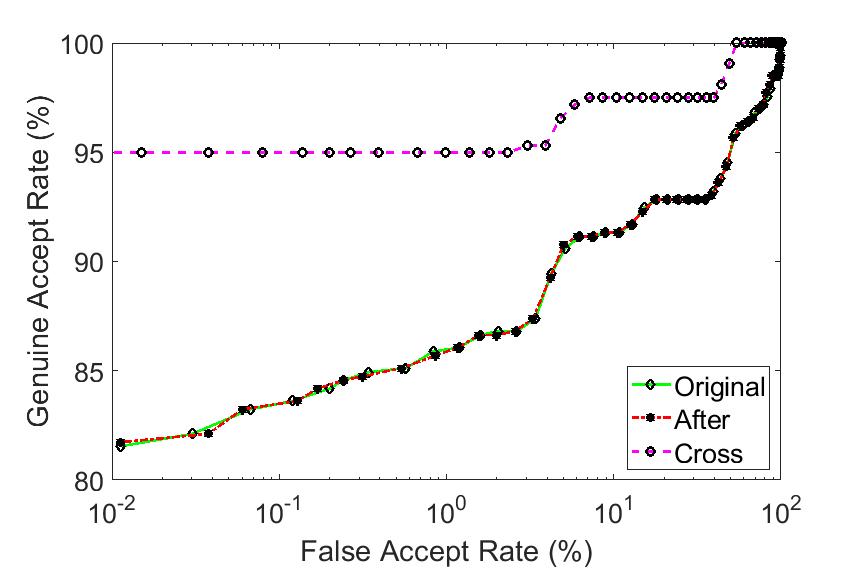} \hspace{-2.9cm}\raisebox{\dimexpr 2.6cm+\height}{\footnotesize{Rear-Indoor}} \hspace{1cm} 
\includegraphics[scale=.13]{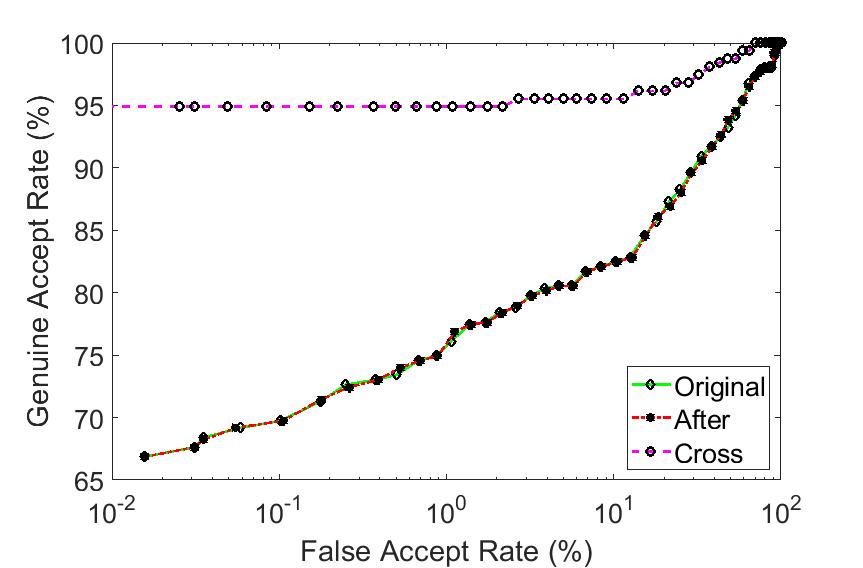} \hspace{-2.9cm}\raisebox{\dimexpr 2.6cm+\height}{\footnotesize{Rear-Outdoor}} \hspace{1cm}
}

\caption{ROC curves for matching \textbf{PRNU Spoofed} images. Here, the source sensor is Device 2 UNIT I. In this case, the target sensors are: (a) Device 1 UNIT I (top row) and (b) Device 1 UNIT II (bottom row).}
\label{Fig: ROC_PRNUspoof3}
\end{figure*}

Next, we report the results for the periocular biometric recognition experiments. The periocular matching experiments indicate the preservation of the biometric utility of the images in both PRNU anonymized images and PRNU spoofed images. The ROC curves corresponding to `Original' and `After' matching experiments are within 1\% of each other. Figure~\ref{Fig: ROC_PRNU Anon} presents the ROC curves for images subjected to PRNU anonymization. Note that Samsung Galaxy S4 results in overall lower periocular matching performance even for the original images.  The cross-matching experiments result in perfect match (100\%) for a majority of the cases barring the Samsung Galaxy S4 sensor. The suppression of the high frequency components may also result in removal of edges and other details which can impact the matching performance. For the PRNU spoofing experiments, we have presented the ROC matching curves for each smartphone device or unit used in this work in Figures~\ref{Fig: ROC_PRNUspoof1},~\ref{Fig: ROC_PRNUspoof2} and~\ref{Fig: ROC_PRNUspoof3}. The matching experiments show that the perturbation scheme used for PRNU spoofing does not degrade the biometric recognition performance. 
 
\section{Conclusion}
\label{sec:concl}

In this work, we design an algorithm that perturbs a face image acquired using a smartphone camera such that (a) sensor-specific details pertaining to the smartphone camera are suppressed (sensor anonymization); (b) the sensor noise pattern of a different device is incorporated (sensor spoofing); and (c) biometric matching using the perturbed image is not affected (biometric utility). We achieve this by applying the Discrete Cosine Transform to images and further modulating the DCT coefficients to either attain PRNU anonymization or PRNU spoofing. In contrast to existing methods which involve computation of sensor reference patterns and exhaustive parameter optimization~\cite{Ref_Fingerprintcopy2, PRNU_attack5_2010, Uhl4_ICB_12}, the proposed method is simple and can achieve highly promising results. In our experiments, we considered face (partial and full) images acquired using the front and rear cameras of different smartphones resulting in data from a total of 12 camera sensors. Our proposed method results in successful camera de-identification for images without compromising the biometric matching performance. An average of $\approx 82.8\%$ reduction in sensor identification is reported in the case of PRNU anonymization, and an average spoof success rate of $\approx 99.5\%$ is observed for PRNU spoofing across the three PRNU estimation schemes evaluated in this work.

\vspace{-0.1cm}
Future work will consider data from a larger number of sensors to evaluate the proposed algorithm. Moreover, the impact of different denoising and PRNU estimation algorithms will be evaluated to study the robustness of the proposed method.  Finally, we will explore methods to make sensor identification schemes resilient to de-identification~\cite{Ross_ICB_19}. 
\balance

{\small
\bibliographystyle{IEEE}
\bibliography{BTAS_REFS}
}
\end{document}